\documentclass[12pt]{spieman}

\usepackage{blindtext}
\usepackage{graphicx}
\usepackage{amssymb}
\usepackage{amsmath}
\usepackage{cite}
\usepackage[parfill]{parskip}
\usepackage[capitalize]{cleveref}
\usepackage{isomath}
\usepackage{upgreek}
\usepackage{ulem}

\newcommand*\xbar[1]{%
   \hbox{%
     \vbox{%
       \hrule height 0.5pt 
       \kern0.3ex
       \hbox{%
         \kern-0.1em
         \ensuremath{#1}%
         \kern-0.1em
       }%
     }%
   }%
} 

\begin{document}
%
\title{Unsupervised Community Detection with a Potts Model Hamiltonian, an Efficient Algorithmic Solution, and Applications in Digital Pathology}
%
%
%

\author[a]{Brendon~Lutnick}
\author[b]{Wen~Dong}
\author[c]{Zohar~Nussinov}
\author[a,d,e,*]{Pinaki~Sarder}
\affil[a]{Department of Pathology and Anatomical Sciences, University at Buffalo $-$ The State University of New York, Buffalo, New York 14203, United~States}
\affil[b]{Department of Computer Science and Engineering, University at Buffalo $-$ The State University of New York, Buffalo, New York 14203, United~States}
\affil[c]{Department of Physics, Washington University in St.\ Louis, St.\ Louis, MO 63130, United~States }
\affil[d]{Department of Biomedical Engineering, University at Buffalo $-$ The State University of New York, Buffalo, New York 14203, United~States}
\affil[e]{Department of Biostatistics, University at Buffalo $-$ The State University of New York, Buffalo, New York 14203, United~States}
\doublespacing
\maketitle
{\footnotesize\textbf{*} Corresponding author, Pinaki~Sarder, Tel : +1-716-829-2265, E-mail : \linkable{pinakisa@buffalo.edu}}

\newpage
\begin{abstract}\label{abstract}
Unsupervised segmentation of large images using a Potts model Hamiltonian~\cite{JMI:JMI12097, doi:10.1117/12.2036875, PhysRevE.81.046114, PhysRevE.85.016101} is unique in that segmentation is governed by a resolution parameter which scales the sensitivity to small clusters. Here, the input image is first modeled as a graph, which is then segmented by minimizing a Hamiltonian cost function defined on the graph and the respective segments. However, there exists no closed form solution of this optimization, and using previous iterative algorithmic solution techniques~\cite{JMI:JMI12097, doi:10.1117/12.2036875, PhysRevE.81.046114, PhysRevE.85.016101}, the problem scales with $(Input Length)^2$. Therefore, while Potts model segmentation gives accurate segmentation~\cite{JMI:JMI12097, doi:10.1117/12.2036875, PhysRevE.81.046114, PhysRevE.85.016101}, it is grossly underutilized as an unsupervised learning technique. We propose a fast statistical down-sampling of input image pixels based on the respective color features, and a new iterative method to minimize the Potts model energy considering pixel to segment relationship. This method is generalizable and can be extended for image pixel texture features as well as spatial features. We demonstrate that this new method is highly efficient, and outperforms existing methods for Potts model based image segmentation. We demonstrate the application of our method in medical microscopy image segmentation; particularly, in segmenting renal glomerular micro-environment in renal pathology. Our method is not limited to image segmentation, and can be extended to any image/data segmentation/clustering task for arbitrary datasets with discrete features. 
\end{abstract}



%

\newpage

\section{Introduction}\label{introduction}
In Potts model based unsupervised segmentation ~\cite{PhysRevE.81.046114}, the image is represented as a graph ~\cite{scheinerman2011fractional, 7322238}. Namely, the pixels are represented by nodes. Relations between the pixels are represented by edges. The energy function (or ``Hamiltonian'')  is adopted from theoretical physics, where it is used to describe ferromagnets and many other systems~\cite{RevModPhys.54.235}. Graph edges are used to update the energy of the partitioned graph, which is iteratively improved until convergence. Unique to this segmentation method is the ability to tune a resolution parameter and modulate the sensitivity to small structures~\cite{Yu2014}. This tuning process allows Potts model based segmentation to indirectly conduct automatic model selection~\cite{PhysRevE.85.016101,castle2010evaluating, PhysRevE.81.046114}.

For segmentation of large images, Hamiltonian minimization quickly becomes computationally unmanageable, and its iterative solution limits parallelization, leading to long run times to reach convergence~\cite{PhysRevE.85.016101, JMI:JMI12097}. The computational challenges here are 2-fold: both graph generation and Hamiltonian minimization for large input datasets have extensive computational overheads. Pixel-scale image segmentation requires an input node for each pixel, and the calculation of a fully connected set of edges (pixel relations). Large edge matrices quickly overwhelm the memory limits of modern hardware, and are intensive to calculate. Hamiltonian minimization has been described as NP-Hard~\cite{Kovtun2003, Liu:2017:SNP:3068585.3068591}. There is no closed form solution for the Potts model Hamiltonian, and existing algorithmic solutions to optimize Potts model based image segmentation are inefficient~\cite{JMI:JMI12097, doi:10.1117/12.2036875, PhysRevE.81.046114}. 

In this paper, we propose a new algorithmic approach for image segmentation that identifies different structures in a medical image, based on the premise that pixels belong to different structures have different local features, including color and texture. Our approach is based on minimizing a Potts model Hamiltonian of the global interactions among the feature-groups in the feature space. This new method differs from many previous methods that minimize the Hamiltonian of the local interactions between neighboring pixels in the image space. It builds on previous implementations to minimize Hamiltonian in the feature space~\cite{JMI:JMI12097, doi:10.1117/12.2036875, PhysRevE.81.046114}, but is able to segment large images exponentially faster through considering the interactions among features groups rather than individual features. Our proposed method in its current set-up can conduct image segmentation based on pixel color information, as well as local texture information. The proposed method can be extended using pixel co-ordinates as features as well. We test the performance of our proposed algorithm quantitatively for segmenting renal histology dataset as a use-case, compare the performance with a dynamic programming mediated optimization based Potts model segmentation method proposed by Storath {\em et al.}~\cite{storath2014fast} and a Markov random field method which uses superpixels for speed gains by Stutz {\em et al.}~\cite{stutz2018superpixels}, and discuss the segmentation performance of these methods in comparison to manual annotations. Our proposed algorithm outperforms the method developed by Storath {\em et al.} and Stutz {\em et al.} in segmenting image regions, while offering slower convergence rate than the latter method. Our segmentation task for pixel color and texture segmentation can also be considered as pixel level data clustering, and therefore we compare the performance of our method with k-means~\cite{1017616, 2016arXiv160202514N, 2016arXiv160202934N, Shindler:2011:FAK:2986459.2986724, Braverman:2011:SKM:2133036.2133039} and spectral clustering~\cite{NIPS2001_2092, 6213172} methods using synthetic data, and obtain comparable or better performance than these classical tools. Our method will find applications in image segmentation tasks involving pixel color and texture clustering, and also in areas (genomics, security, and social media) involving data clustering with each data-point representing a set of multivariate features. 

This paper is organized as follows. In \cref{method}, we describe the proposed algorithm for Potts model energy minimization. In \cref{results}, we present the performance of the proposed method using real images of renal tissue histology and synthetic data. In \cref{discussion}, we discuss the results presented in \cref{results}, and conclude in \cref{conclusion}. 

\section{Method}\label{method}
Hu {\em et al.}~\cite{PhysRevE.85.016101,JMI:JMI12097} have implemented a Potts model minimization to segment medical and other images with high accuracy. These works investigate the full graph and minimize the Potts model using an approach relying on ``trials'' and ``replicas''; in these works, the energy is minimized in an iterative fashion. Namely, starting from an initial segmentation, the algorithm is allowed to converge to a local solution for multiple iteration moves (trials) and the Potts configuration that best minimizes the energy is chosen. Collectively, an ensemble of lowest energy candidate solutions found amongst several trials (this ensemble is that of ``replicas'') allows, via a calculation of information theory metrics, an automated inference of pertinent structures and the optimal parameters appearing in the Potts model Hamiltonian (including, notably, the resolution parameter $\gamma$ that we will introduce in Section \ref{graph segmentation}). During each energy minimization move, the algorithm searches for a segment to which a given node may be assigned to such that this assignment lowers the energy. Albeit providing very accurate segmentations, this method is limited in its performance due to its slow convergence rate; this method exhaustively considers the full graph and investigates, in some detail, node-node relationships during the search moves. Such a modus operandi is indeed somewhat inefficient. Indeed, myriad medical image segmentation problems can be solved via color and texture clustering of the pixels. With this in mind, we developed a solution of Potts model energy minimization that considers selected nodes of the full graph by exploiting the fact that pixels in a given segment in a medical image typically show similar color and texture. During the energy minimization, for a given node, instead of comparing the node with all the other nodes, we furthermore consider placing this node in all the other segments to determine an energy lowering move. This process greatly increases the segmentation efficiency while offering high accuracy. Proposed segmentation steps are discussed below.  

\subsection{Image as a Graph}\label{dataset definition}
A 3-channel $RGB$ image of pixel size $k \times l$ is represented as a vectorized data $\mathbf{D}$ of length $M = k \times l$ (i.e., $M$ is the number of pixels in the image) with $n=3$ associated color features: 
\begin{equation}\label{eq1}
\mathbf{D} = 
	\left(\begin{array}{cccc}
	d_{11} & d_{12} & d_{13} \\
	d_{21} & d_{22} & d_{23} \\
	\vdots & \vdots & \vdots \\
	d_{M1} & d_{M2} & d_{M3}
	\end{array}\right).
\end{equation}
The number of features along the columns can further be appended using local pixel textures as well. For the sake of clarity, we discuss our algorithm using (three) pixel color information only. Here, the image segmentation is viewed as a clustering of the rows of $\mathbf{D}$, which is equivalent to image segmentation based on pixel color information. The output of the algorithm is a vector of length $M$ with each element containing a segmentation index of a respective pixel of the original image. 

In order to perform image segmentation, the image data $\mathbf{D}$ is converted to a graph. The rows of $\mathbf{D}$ describe the nodes; the relationships (e.g., Euclidean distances between two rows computed based on the respective features) between the rows are detailed as edges in the graph. In general, it is not efficient to employ all the rows of $\mathbf{D}$ as nodes. To that end, we use a highly efficient node selection process as described below for our proposed Potts model energy minimization study. The node selection process first excludes the redundant data, and then performs a down-sampling operation to simplify the graph. In this way, each row or data-point defined in $\mathbf{D}$ is associated with a node. This process lowers the effective size of the graph, ensuring the graph is computationally manageable, essentially performing an over-segmentation~\cite{DBLP:journals/corr/AmorimMM16} of the original image.

The image color information in $\mathbf{D}$ typically contains several rows with similar color values. Therefore, to more efficiently conduct an image segmentation based on color, we will automatically group together individual pixels having identical colors and only analyze unique rows in $\mathbf{D}$. For the purpose of labeling the RGB values of individual pixels, any mathematical operation on the rows of $\mathbf{D}$ can be used. In this work, we chose to use a Cantor pairing operation~\cite{Hopcroft:2006:IAT:1196416, Cantor}. This operation produces a unique number corresponding to each feature vector or row in $\mathbf{D}$. The Cantor pairing output between the first two features in the $i^{\mathrm{th}}$ row of $\mathbf{D}$ is given as:
\begin{equation}\label{eq6}
\pi_i(d_{i1}, d_{i2}) = \frac{1}{2}(d_{i1} + d_{i2})(d_{i1} + d_{i2} +1) + d_{i2},  
\end{equation} 
where $i \in \left\{1,2,...,M\right\}$. Next, the Cantor pairing between $\pi_i(d_{i1}, d_{i2})$ and $d_{i3}$ is computed. For the $n-$feature case, this process is subsequently repeated for $n-1$ times for each row of $\mathbf{D}$. Unique values of the resulting Cantor pairing outputs corresponding to the rows of $\mathbf{D}$ identify specific colors in the original image. Using this algorithmic step, we thus define a reduced $\tilde{\mathbf{D}}$ using the unique image colors as, 
\begin{equation}\label{eq8}
\tilde{\mathbf{D}} = 
	\left(\begin{array}{cccc}
		\tilde{d}_{11} & \tilde{d}_{12} & \tilde{d}_{13} \\
		\tilde{d}_{21} & \tilde{d}_{22} & \tilde{d}_{23} \\
		\vdots & \vdots & \vdots \\
		\tilde{d}_{M'1} & \tilde{d}_{M'2} & \tilde{d}_{M'3}
	\end{array}\right),
\end{equation}
where $\tilde{d}_{ij}$ are the features of the reduced dataset, containing $M' \le M$ data-points. Each row in $\tilde{\mathbf{D}}$ is different from all other rows and has a different Cantor pairing output. Note that the rows in $\tilde{\mathbf{D}}$ are essentially a subset of the rows defined in $\mathbf{D}$ corresponding to the original image. The correspondence between the rows in $\mathbf{D}$ and $\tilde{\mathbf{D}}$ is kept stored in the computer, information of which we use later to form the output segments corresponding to the input image scale using the Potts model minimization output obtained from the reduced number of nodes. Essentially $\tilde{\mathbf{D}}$ can be used to form a graph corresponding to the image color information using the rows as the nodes and relationships between the rows as the respective edges between the nodes following the similar discussion as before. Reducing $\mathbf{D}$ to its distinct entries via the mapping $\mathbf{D} \to \tilde{\mathbf{D}}$ forces the resulting Potts model segmentation to scale with the feature-space volume rather than the length of $\mathbf{D}$. For a dataset with discrete bounded features (pixel R, G, B values), this method often represents a significant reduction in the considered data, and thus improves the speed of Potts model minimization.  

To further enhance the minimization speed, statistical down sampling can be used to reduce further the length of $\tilde{\mathbf{D}}$. We apply this operation if the length of $\tilde{\mathbf{D}}$ is greater than a user defined threshold (set to $500$ in our simulations). The goal is to group pixels of similar color values. Towards that end, we use a modified K-means algorithm with a user defined $K$, reducing the length of the $\tilde{\mathbf{D}}$ to be $K$. As we will explain below, this leads to a new matrix 
$\tilde{\mathbf{D}}'$ before constructing the graph for the Potts model minimization. We apply the K-means algorithm after a prior Cantor pairing; such a sequence of operations improves the speed of the K-means for color based segmentation of very large images (typically medical images, e.g., histology images, have a limited color palette). The $K$ is chosen to be larger than the expected number of segments present in an image, allowing faster segmentation while introducing negligible error. 

Because the precise value of $K$ is not critical, we use a modified K-means algorithm optimized for speed for the statistical down sampling. Rather than employing a traditional K-means minimization~\cite{1017616}, we perform K-means clustering individually for each column of 
$\tilde{\mathbf{D}}$ so as to partition each feature dimension independently. This method surveys the feature-space rather than the image data-structure to determine revised $\approx K$ data-points pertaining to image color. Using this method, the number of groups ($k_1,$ $k_2,$ and $k_3$) for the three columns of $\tilde{\mathbf{D}}$ is set to be
\begin{equation}\label{eq12}
k_j = \left\lfloor  \frac{\sigma^2_j}{ \sum_{j} \sigma^2_j } \left( \frac{K}{\left( \prod_j \sigma^2_j \right)^{1/3}} \right) + \frac{1}{2} \right\rceil.
\end{equation}
Here, $j\in\{1,2,3\}$, $\sigma^2_{j}$ is the variance of the elements in the $j^{\mathrm{th}}$ column of $\tilde{\mathbf{D}}$ and $\left\lfloor \bullet \right\rceil$ is the rounding operation. 

A K-means clustering~\cite{1017616} is performed independently on each column of $\tilde{\mathbf{D}}.$ Recall that $\tilde{\mathbf{D}}'$ is obtained via this modified K-means algorithm discussed herein, reducing the length of the $\tilde{\mathbf{D}}$ to be $K.$ To construct $\tilde{\mathbf{D}}'$, each element of the $j^{\mathrm{th}}$ column of $\tilde{\mathbf{D}}$ is first replaced by the corresponding mean of the K-means clustered partition that this element belongs to. Then another Cantor pairing operation row-wise is applied to the resulting matrix, as before with a goal to obtain unique rows, to generate $\tilde{\mathbf{D}}'$. This technique provides a non-linear down-sampling of the input data, which we have found to converge significantly faster than traditional K-means clustering~\cite{1017616}. This technique however does not guarantee always $K$ rows in the resulting $\tilde{\mathbf{D}}'$.

If the user prefers the length of $\tilde{\mathbf{D}}'$ to be $K,$ and this is not achieved in the previous step, then $k_1,$ $k_2,$ $k_3,$ as well as $\tilde{\mathbf{D}}'$ are iteratively updated. Using the respective values of these variables and matrix as obtained in the previous paragraph as initial values, we use a discrete version of proportional-integral-derivative (PID) control system~\cite{1453566, 6341788} to iteratively update $k_1,$ $k_2,$ $k_3,$ and $\tilde{\mathbf{D}}'.$ This process is repeated until the length of $\tilde{\mathbf{D}}'$ becomes at least $\alpha K,$ where $\alpha$ is a user defined constant. Further discussion appears in \cref{discussion}. Following the above steps, the algorithm quickly renders the length of $\tilde{\mathbf{D}}'$ to be the desired 
$\approx K$ in a generalizable fashion. The values of $k_1,$ $k_2,$ and $k_3$ at each iterative step are adjusted according to
\begin{equation}\label{eq18}
k_j^{(t)} = \left\lfloor k_{j}^{(t-1)}\left( K_\mathrm{p} e^{(t)} + K_\mathrm{i} \sum_{i=0}^{t} e^{(i)} + K_\mathrm{d} (e^{(t)}-e^{(t-1)}) + 1 \right) \right\rceil.
\end{equation}
Here $t$ labels the iteration step. The error $e^{(t)}$ at this iteration is given by
\begin{equation} \label{eq19}
e^{(t)} = 1 - \frac{length(\tilde{\mathbf{D}}'^{(t)})}{K}.
\end{equation} 
A discussion of the tuning parameters $K_\mathrm{p}$, $K_\mathrm{i}$, and $K_\mathrm{d}$ can be found in \cref{discussion}. 
The correspondence between the rows in $\tilde{\mathbf{D}}$ and $\tilde{\mathbf{D}}'$ is stored in the computer as before. This information will later allow us to form the output segments corresponding to the input image scale using the Potts model minimization output obtained using the reduced number of nodes of $\tilde{\mathbf{D}}'$.

The graph is defined using the rows of $\tilde{\mathbf{D}}'$ as nodes. Assume the length of $\tilde{\mathbf{D}}'$ to be $M''$ (a number close to $K$). The Euclidean distance between any pair of nodes $i^{\mathrm{th}}$ and $j^{\mathrm{th}}$ with $\forall i,j\in\{1,2,\dots,M''\}$ defines the value of the edge ($e_{ij}$) connecting these two nodes. Edges are calculated for all combinations of nodes. Assume that the total number of edges are $N$. The average edge value is required as the background in the Potts model minimization discussed below. Since the matrix $\tilde{\mathbf{D}}'$ is formed via several levels of reduction of the original image data matrix $\mathbf{D}$, the mean of the edges of the graph formed based on $\tilde{\mathbf{D}}'$ is not representative of the true edge average of the graph corresponding to the image data $\mathbf{D}$. Therefore computing the average edge value from 
$\mathbf{D}$ is recommended. However, because of the large size of the original image, such computation can be intensive. To avoid calculating edges for the full data set, in this work, we employ a uniform down-sampling of the data, reducing it to a maximum length of $M''',$ where $M'' \leq M''' \leq M$ is satisfied. A full set of edges is computed using this reduced data-points where graph edges are calculated using a similar procedure as described above. Average edge value is computed to be $\bar{e}$. Given $M'''$ is sufficiently large, $\bar{e}$ will asymptotically approximate the true average edge closely, while greatly reducing the algorithmic overhead. Further discussion on $M'''$ can be found in \cref{discussion}.

\subsection{Graph Segmentation}\label{graph segmentation}
The graph is segmented by minimizing a modified Potts model Hamiltonian~\cite{PhysRevE.81.046114}. A detailed explanation of Potts model minimization based image segmentation is discussed in our earlier works; see works by Hu {\em et al.}~\cite{PhysRevE.85.016101,JMI:JMI12097}. The Potts model Hamiltonian of the graph corresponding to $\tilde{\mathbf{D}}'$ is given by,
\begin{equation} \label{eq25}
\mathcal{H} = \sum_{j=i+1}^{M''} \sum_{i=1}^{M''} (e_{ij} - \xbar{e}) \left[\Theta (\xbar{e} - e_{ij}) + \gamma \Theta (e_{ij} - \xbar{e}) \right] \delta(S_{i},S_{j}).
\end{equation} 
The Heaviside function \cite{weisstein2} determining which edges are considered is given by
\begin{equation}\label{eq26} 
\Theta (e_{ij} - \xbar{e}) = \left\{
        \begin{array}{ll}
            1, & \quad e_{ij} > \xbar{e}, \\
            0, & \quad \mathrm{otherwise.}
        \end{array}
    \right.
\end{equation} 
The resolution parameter $\gamma$ is used to tune the segmentation. This user selected parameter determines the number of segments to be obtained by minimizing $\mathcal{H}.$ Decreasing $\gamma$ results in segments with lower intra-community density, revealing larger communities, or lower number of segments. Increasing $\gamma$ results in smaller communities, or a higher number of segments. In our previous work~\cite{PhysRevE.85.016101}, we have also shown that $e_{ij}$ can be modulated using the distance between the respective ($i^{\mathrm{th}}$ and $j^{\mathrm{th}}$) pixels, and this modulation indirectly controls the estimated segment sizes. The Kronecker delta \cite{weisstein} is defined by
\begin{equation} \label{eq27}
\delta(S_{i}, S_{j}) = \left\{
        \begin{array}{ll}
            1, & \quad S_{i} = S_{j}, \\
            0, & \quad \mathrm{otherwise.}
        \end{array}
    \right.
\end{equation}
In Eq. (\ref{eq25}), the Kronecker delta $\delta(S_i, S_j)$ ensures that each node (spin) interacts only with nodes in its own segment. Here, $S_i$ denotes the segment to which the $i^{\text{th}}$ node belongs to. The segment identities are determined by minimizing the Hamiltonian energy $\mathcal{H}$, thus, giving the segmented graph $S=\{S_1,S_2,\dots,S_{M''}\},$ where $S_i$ is an integer and $i \in \{1,2,\dots,L\}$, 
with $L$ being total number of segments. 

There exists no closed form solution for the lowest energy states of the Hamiltonian \cref{eq25}. Moreover, the number of the possible solutions to the segmentation problem scales exponentially in the total number ($N$) of edges in the graph to be segmented. In addition to recasting the image segmentation problem as that associated with a smaller size color based image graph for which segmentation is more efficient (as we discussed above), our major contribution in minimizing the Potts model Hamiltonian is a new algorithmic approach that we detail below. By comparison to the previous approach of Hu {\em et al.} and others~\cite{JMI:JMI12097, doi:10.1117/12.2036875, PhysRevE.81.046114, PhysRevE.85.016101}, our proposed optimization reduces both the computational overhead as well as the number of (user specified or other) parameters employed. 

\begin{figure}
  \includegraphics[width=\linewidth]{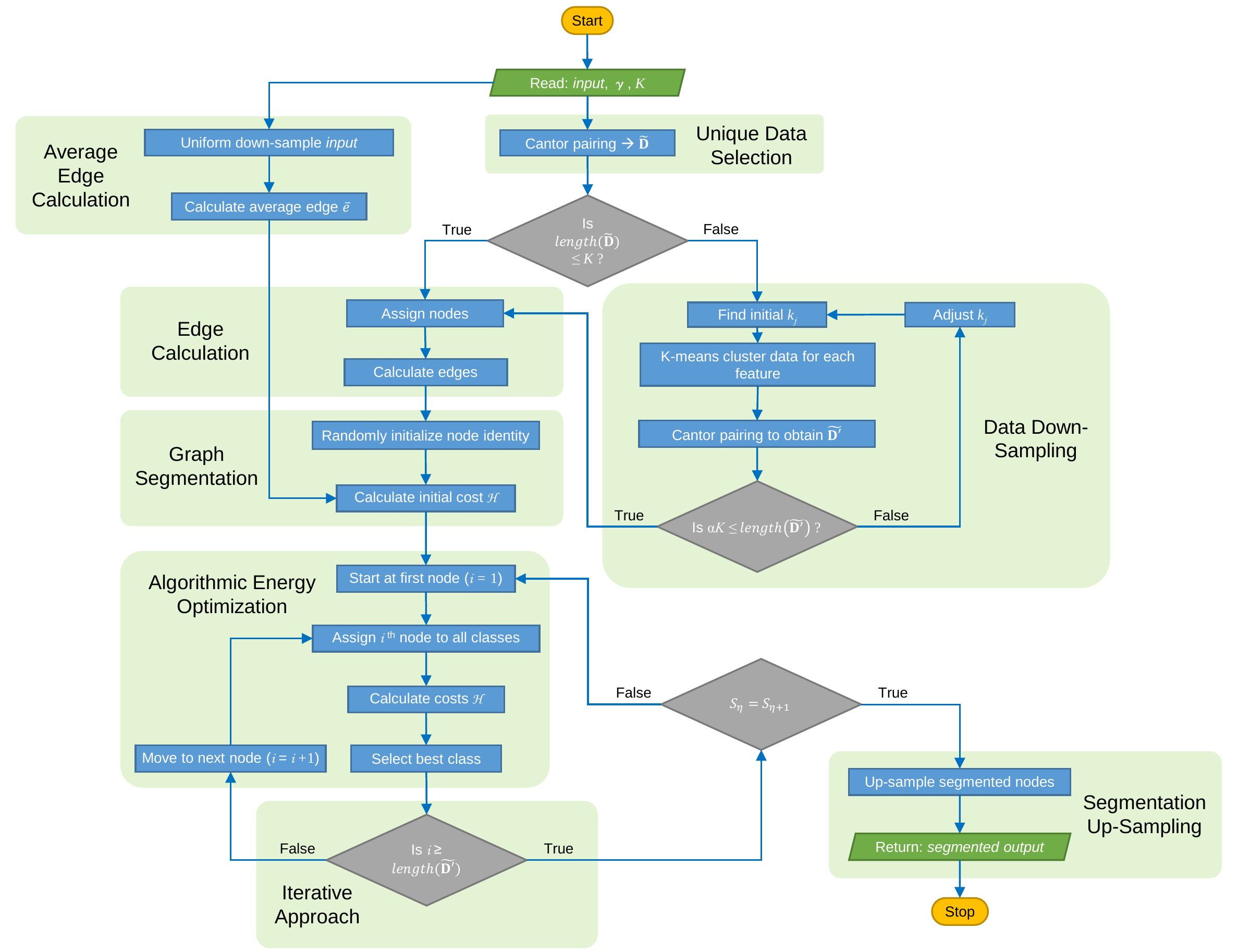}
  \caption{Algorithm flow chart detailing our iterative solution pipeline.}
  \label{fig:flowchart}
\end{figure}

In previous works~\cite{JMI:JMI12097, doi:10.1117/12.2036875, PhysRevE.81.046114, PhysRevE.85.016101}, a gradient descent type minimization of $\mathcal{H}$ was performed starting from an initial seed state. This initialization was one of two types. One approach was to initially have each of the nodes constitute a different individual segment (i.e., the number of segments was equal to the number of nodes). A second approach was to randomly initialize the starting node identities into a fixed number of segments larger than the optimal $L$. During the minimization, the energy associated with moving each node to the segment of a different node (i.e., allowing it to a fuse with the segment to which this different node belongs to) was computed to determine if such a reassignment of the node will lower  $\mathcal{H}$; the comparison of the energy (to ascertain if moving the node is energetically preferable) was done by considering node pairs. Thus, as the minimization proceeded, the number of segments decreased monotonically in time. Energy lowering moves were applied to all nodes; the process was repeated until no further energy lowering moves were found in subsequent iterations. The outcome of this process was a candidate low energy state- a ``solution''. While accurate, this way of minimizing $\mathcal{H}$ allowed for multiple solutions when following different stochastic attempts (trials) to minimize the Hamiltonian from the same start-point. Therefore, earlier work selected the solution with minimal energy after conducting several trials. Due to its implementation, this inherently led to a lower number of segments than those in the initial state. In the current work, we propose a new modification to this strategy. We first randomly assign the initial number of segments for the nodes $\{1,2,\dots,M''\}$ in the reduced graph in a similar way as discussed above. However, during energy lowering moves, each node is either assigned to be the member of each of the other segments or as an individual segment, and the solution corresponding to the assignment that provides the minimal energy is chosen. The process is repeated in a complete iteration, and continued until no energy lowering move is made in two subsequent iterations. This implementation endows the algorithm more freedom to expand or reduce the number of segments as the system ``evolves'' from the initial state. That is, the number of segments is no longer monotonic in the run time. More importantly, we may analyze node-segment relationship to make an energy lowering move and thus parallel implementation is feasible. Overall this new implementation of Potts model minimization is more efficient than previous implementations~\cite{JMI:JMI12097, doi:10.1117/12.2036875, PhysRevE.81.046114, PhysRevE.85.016101}. Previous works considered a replica approach~\cite{JMI:JMI12097, doi:10.1117/12.2036875, PhysRevE.81.046114, PhysRevE.85.016101}; namely, we studied the image segmentation for different randomly selected start-points. For the sake of brevity, we refrain from this analysis in this current work. The optimally segmented graph is finally up-sampled, reversing the Cator pairing and down-sampling performed in \cref{dataset definition}, to determine the segmented dataset. An overview of the algorithmic pipeline and iterative solution is detailed in \cref{fig:flowchart}.

\subsection{Existing Literature}\label{literature}


Various approaches have been developed to segment an image based on Potts model, where pixels within the same segment have similar visual features, and pixels belonging to adjacent segments have significantly different visual features. Most existing approaches work in the image space. With Potts (Mumford-Shah) model, image segmentation is through approximating the original image with a piece-wise constant image that has minimum total boundary length of the pieces~\cite{mumford1989optimal,storath2014fast}. With Markov random field, image segmentation is reduced to maximizing the posterior probability of segment assignment of the pixels with respect to a set of features associated to these pixels and the prior knowledge of how pixels with segment assignments interact with one another \cite{geman1993stochastic,boykov_graphcut,mori2005guiding}. 

Our approach differs from the other approaches in that it segments an image in the feature space, where the interaction between two points in the feature space is defined in terms of their Euclidean distance and whether they belong to the same segment (Eq. \ref{eq25}). In other words, it minimizes Potts model  Hamiltonian (Eq. \ref{eq25}) defined on the complete graph of features in the features space, while many other approaches only consider the local interactions between neighboring pixels in the image space in minimizing the Hamiltonian. Our approach best reflects the needs in medical image segmentation because here our main interest is to identify the different structures (e.g., nuclei, mesangial matrix and Bowman's space in a glomerulus image) based on the premise that they have different local features, including color and texture. This approach builds on the previous works of Hu et al.~\cite{JMI:JMI12097, doi:10.1117/12.2036875, PhysRevE.81.046114, PhysRevE.85.016101}, and significantly improves the segmentation speed by applying vector quantization to divide a large set of features into a small number of super-features (groups) each having the same number of feature points closest to them (Section \ref{dataset definition}), and then minimizes the Hamiltonian involving the global interactions among these super-features (Section \ref{graph segmentation}). This approach differs from the naive approaches of directly applying k-means and spectral clustering algorithms to the features in the feature space in the usage of Potts model Hamiltonian.

\section{Results}\label{results}
Our optimized Potts model segmentation method was evaluated for speed and segmentation performance using benchmark images~\cite{MartinFTM01}, as well as segmentation of histologically stained brightfield microscopy images of murine glomeruli. For the latter case, we compare the performance of our method with the one proposed by Storath {\em et al.}~\cite{storath2014fast}, and a Markov random field method which uses superpixels for speed gains by Stutz {\em et al.}~\cite{stutz2018superpixels}. For simplicity, we use image color (R, G, \& B values) as image features. Additionally, a synthetic dataset was used for a more rigorous quantitative validation of the method. Here, we compare the performance of our method against two other unsupervised segmentation methods: spectral clustering~\cite{NIPS2001_2092, 6213172} and K-means clustering~\cite{1017616, 2016arXiv160202514N, 2016arXiv160202934N, Shindler:2011:FAK:2986459.2986724, Braverman:2011:SKM:2133036.2133039}. 

\subsection{Emperical Comparison with Optimization by Hu {\em et al.}~\cite{JMI:JMI12097, doi:10.1117/12.2036875, PhysRevE.81.046114, PhysRevE.85.016101}}\label{emphu}
\begin{figure}
  \includegraphics[width=\linewidth]{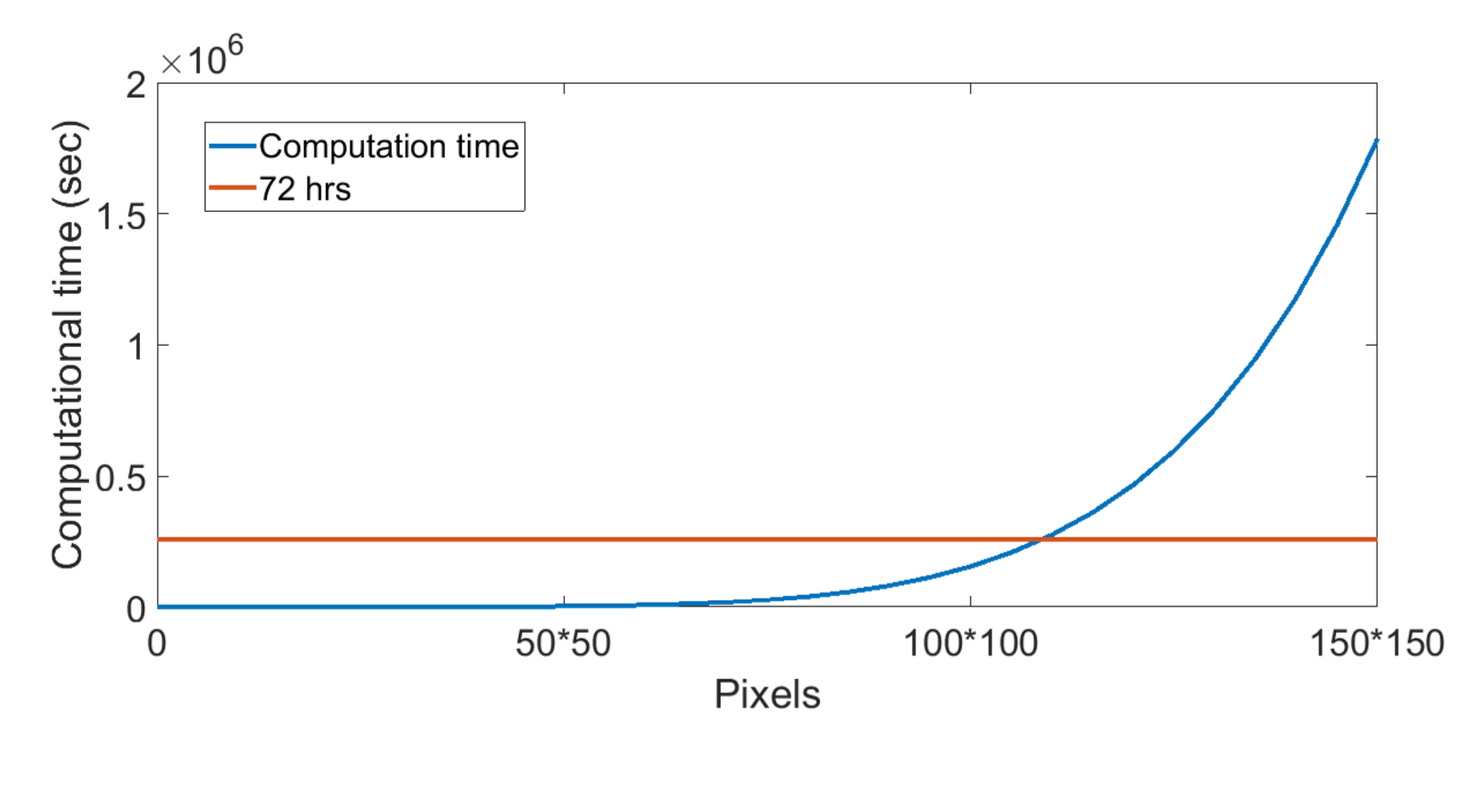}
  \caption{Computational complexity of full graph segmentation using optimzation method proposed by Hu {\em et al.}~\cite{JMI:JMI12097, doi:10.1117/12.2036875, PhysRevE.81.046114, PhysRevE.85.016101}.}
  \label{fig:emphu}
\end{figure}

In \cref{fig:emphu}, we plot the computational complexity of the optimization method developed by Hu {\em et al.}~\cite{JMI:JMI12097, doi:10.1117/12.2036875, PhysRevE.81.046114, PhysRevE.85.016101} for a full graph segmentation. Computational complexity computation here assumes no parallelization. We see that the computational time becomes quickly unmanageable for images of size slightly larger than $100 \times 100$ pixels; for images larger than this size, the images need to be cropped in overlapping blocks and subsequently processed in parallel. Our proposed method in the current work can segment high resolution images of size $~500 \times 500$ in ~2 mins. The brightfield microscopy images of glomeruli have similar sizes, processing of which is discussed below.   

\subsection{Synthetic Data Segmentation}\label{synthetic data segmentation}
To quantitatively assess our method, a synthetic dataset was generated, and segmentation performance was evaluated using information theoretic measures~\cite{bouma2009normalized, 6415998}. 

\subsubsection{Synthetic Data Generation}\label{synthetic data generation}
\begin{figure}
  \includegraphics[width=\linewidth]{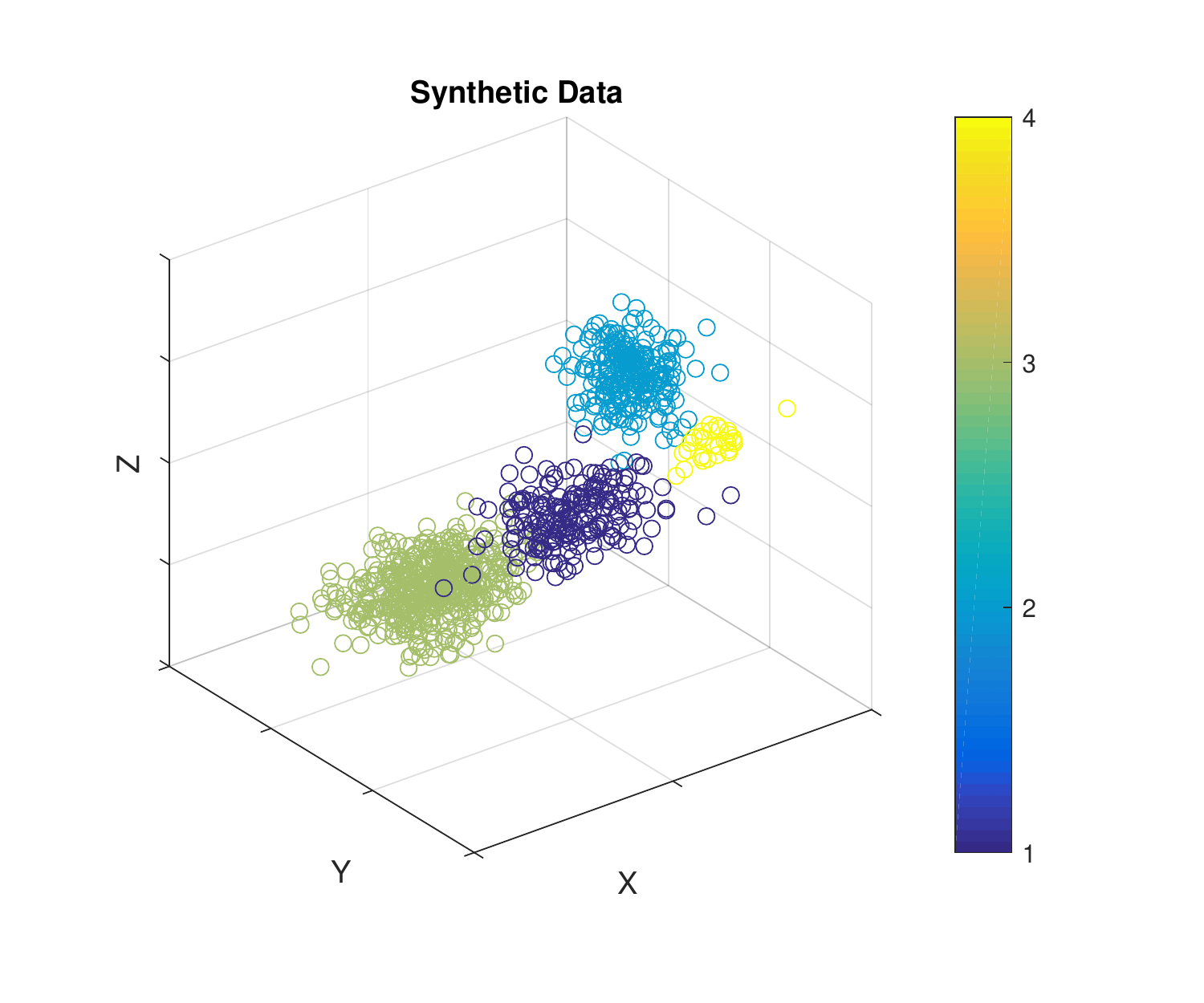}
  \caption{Synthetic Data, containing 1000 points in 4 normally distributed classes.}
  \label{fig:synthdata}
\end{figure}

To ensure that the synthetic dataset was easily separable in its given feature space, clusters were defined as 3-dimensional Gaussian distributions with dimension independent mean and variance. Altering the x, y, and z mean and variance for each distribution controlled the separability of the clusters in the feature space. The number of nodes in each cluster was also altered. An example of this synthetic data is given in \cref{fig:synthdata}. For evaluation, the mean and variance values of the synthetic distributions were changed periodically to ensure robustness. However, all datasets were designed to give a small amount of overlap between classes. 

\subsubsection{Evaluation Metric}\label{evaluation metric}
To quantitatively evaluate clustering performance on the synthetic data, we utilized information theoretic measures~\cite{bouma2009normalized, 6415998}. Specifically, the performance of each method was evaluated by calculating the Normalized Mutual Information (NMI = $I_{\mathrm{N}}$) between the clustered data $c$, and ground truth labels $g$,  
\begin{equation}\label{eq33}
I_{\mathrm{N}}(c,g) = \frac{2I(c,g)}{H_{c}+H_{g}},
\end{equation}
where $0 \leq I_{\mathrm{N}} \leq 1$. Here $H$ is the Shannon entropy, and $I(c,g)$ is the mutual information between $c$ and $g$. These metrics are given as:
\begin{equation}\label{eq34}
H_{c} = -\sum^{L_c}_{k=1}{\frac{N_{k}}{M}\log_{2}\frac{N_{k}}{M}  }
\end{equation}
and
\begin{equation}\label{eq35}
I(c,g) = \sum^{L_{c}}_{k_{1} = 1}{  \sum^{L_{g}}_{k_{2}=1}{  \frac{N_{k_{1}k_{2}}}{M} \log_{2}\frac{N_{k_{1}k_{2}}M}{N_{k_{1}}N_{k_{2}}}  }}.
\end{equation}
Here, $N_{k}$ represents the cardinality of the $k^{\text{th}}$ segment (i.e., the number of pixels in that segment). Likewise, $N_{k_1 k_2}$ denotes the common pixels in the $k_1^{\mathrm{th}}$ segment of $c$ and $k_2^{\mathrm{th}}$ segment of $g$, and $M$ is the total number of data-points. 

\subsubsection{Potts Model Performance}\label{potts model performance}
\begin{figure}
  \includegraphics[width=\linewidth]{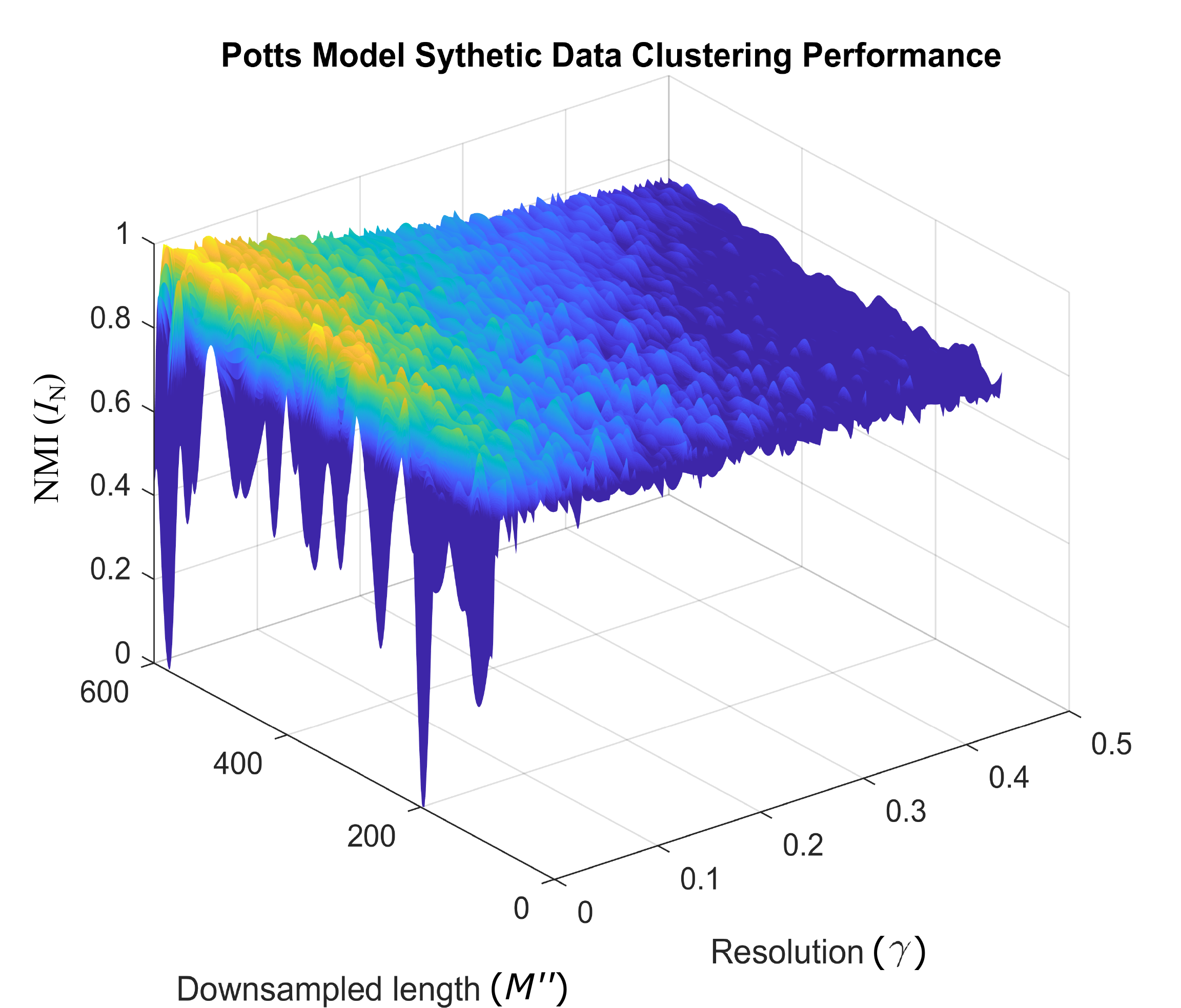}
  \caption{$I_{\mathrm{N}}(\gamma, M'')$ - Potts model performance (NMI) as a function of resolution and nodes, when clustering synthetic data (shown in \cref{fig:synthdata}). Due to the simple nature of the dataset clustered, we observe optimal performance at a low resolution $\gamma = 0.02$. While the clustering performance increases with the number of included nodes, $M''$, near optimal performance is observed when $M'' \geq 350$. }
  \label{fig:surfNMI}
\end{figure}

\begin{figure}
  \includegraphics[width=\linewidth]{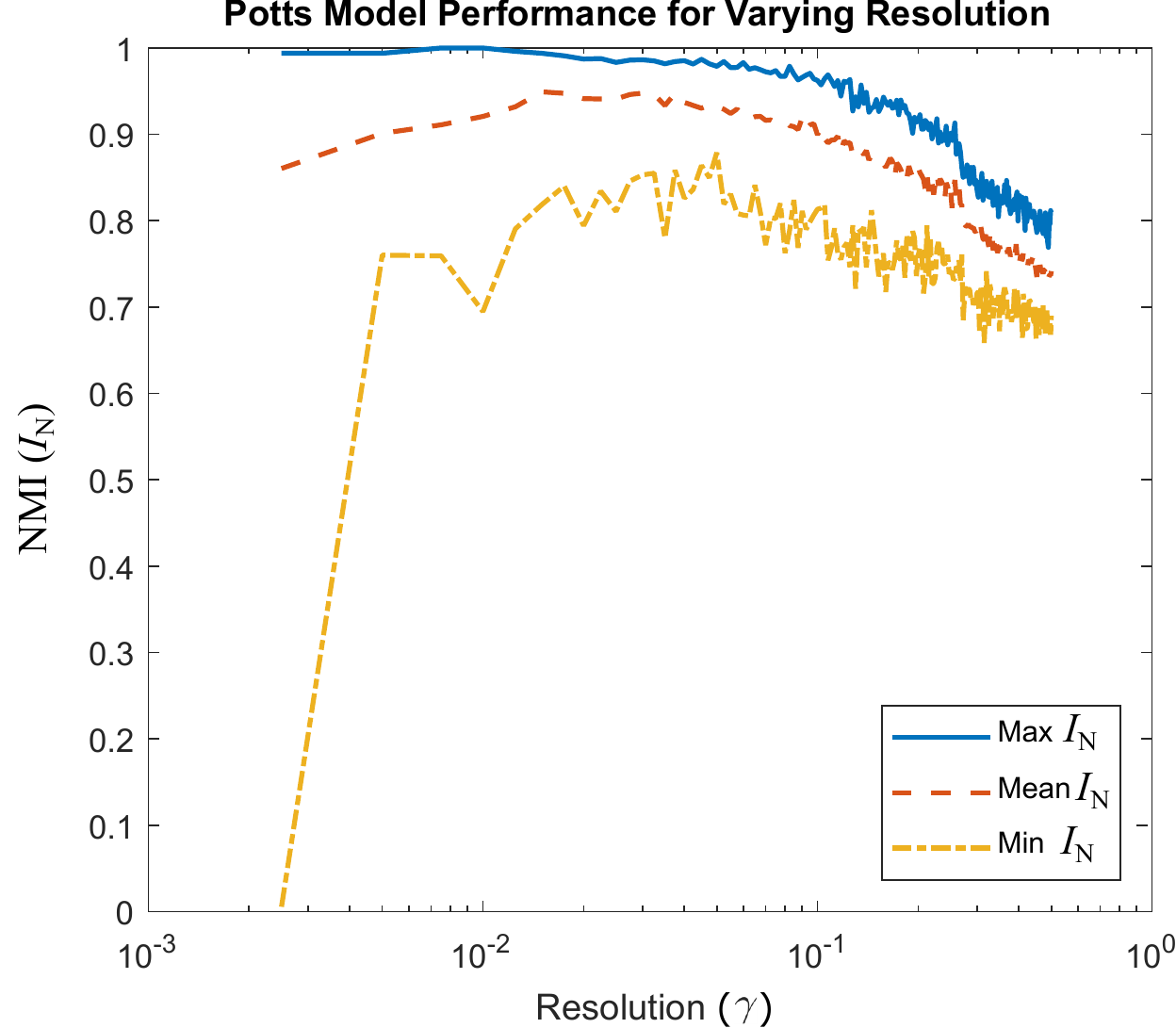}
  \caption{Potts model performance as a function of resolution $\gamma$, when clustering synthetic data (shown in \cref{fig:synthdata}). The best performance is achieved at $\gamma=0.02$. The result represents 36 realizations at each resolution, varying $M''$ between $350 \leq M'' \leq 600$. We observe optimal performance at $\gamma = 0.02$.}
  \label{fig:NMIvsgamma}
\end{figure}

\begin{figure}
  \includegraphics[width=\linewidth]{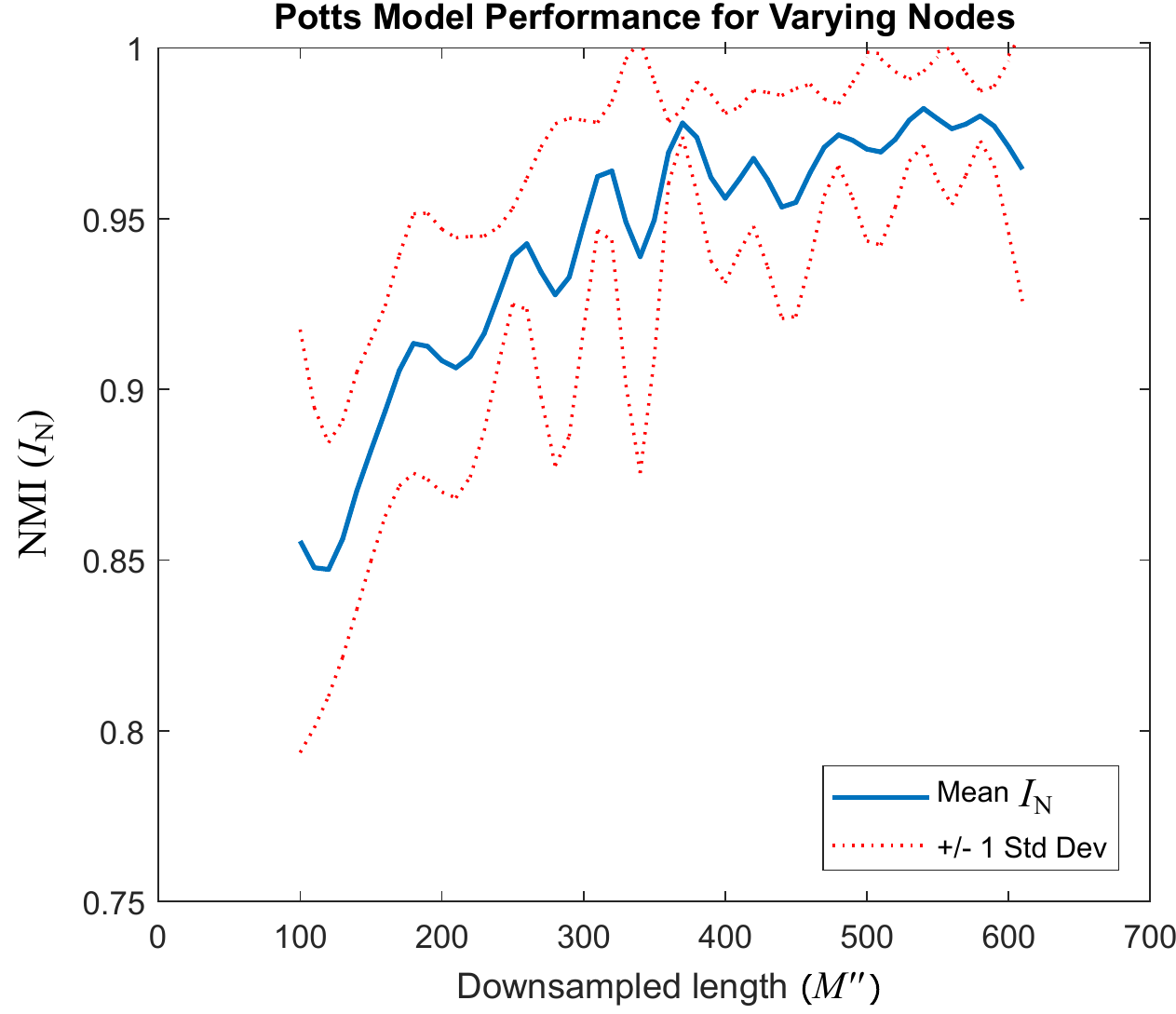}
  \caption{Average Potts model performance as a function of the number of nodes $M''$, when clustering synthetic data (shown in \cref{fig:synthdata}). Clustering at $0.01\leq\gamma\leq0.03$ are included in the averaged performance. The result represents 15 realizations at each down-sampled length. We observe that clustering performance is consistent for $350 \leq M'' \leq 600$.}
  \label{fig:NMIvsphi}
\end{figure}

\begin{figure}
  \includegraphics[width=\linewidth]{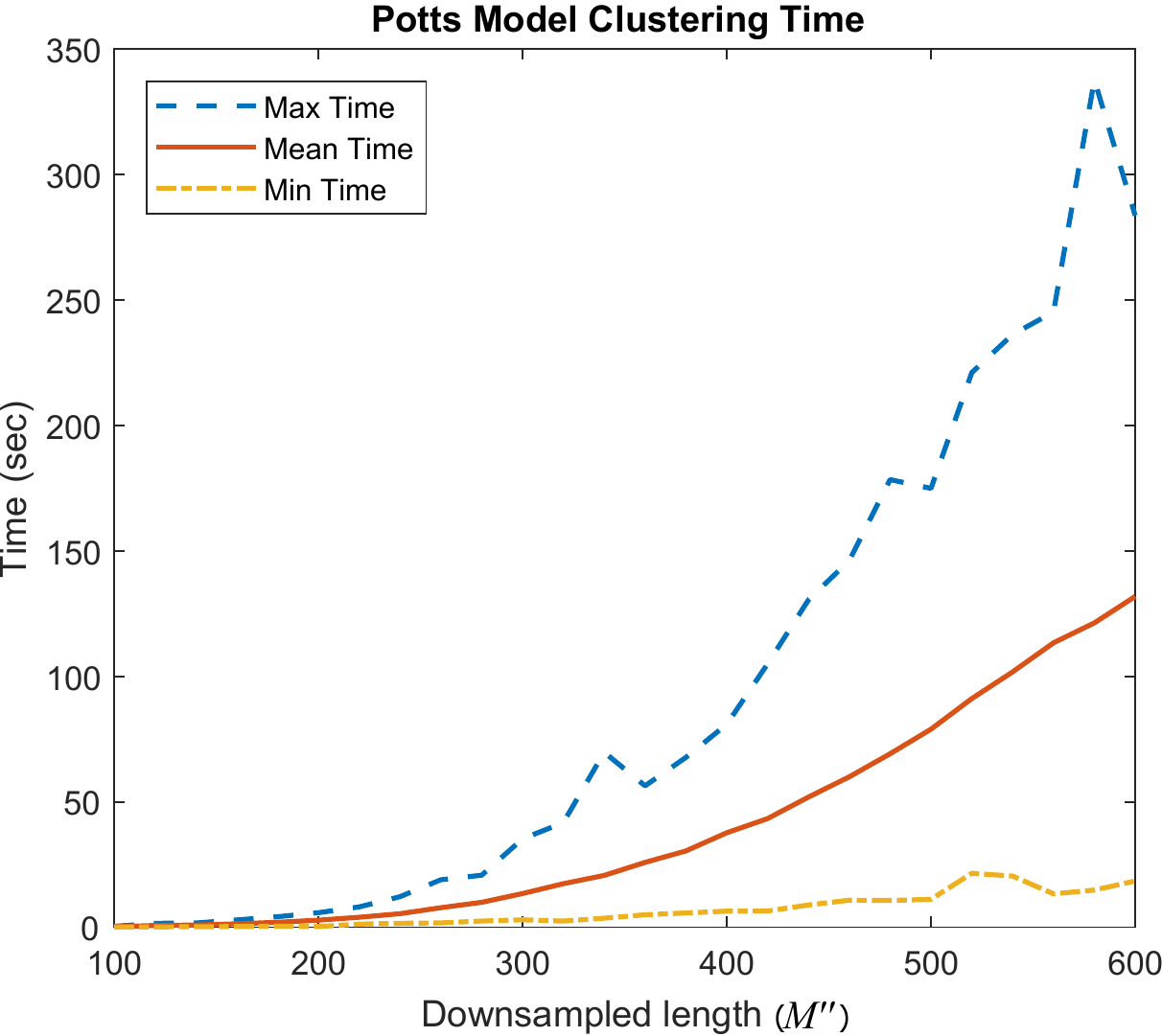}
  \caption{Potts model convergence times as a function of the number of nodes $M''$, when clustering synthetic data (shown in \cref{fig:synthdata}). We observe that our algorithm scales with $\approx M''^2 - M''$ as expected, highlighting the effect of reducing $M''$.}
  \label{fig:timevsnodes}
\end{figure}

\begin{figure}
  \includegraphics[width=\linewidth]{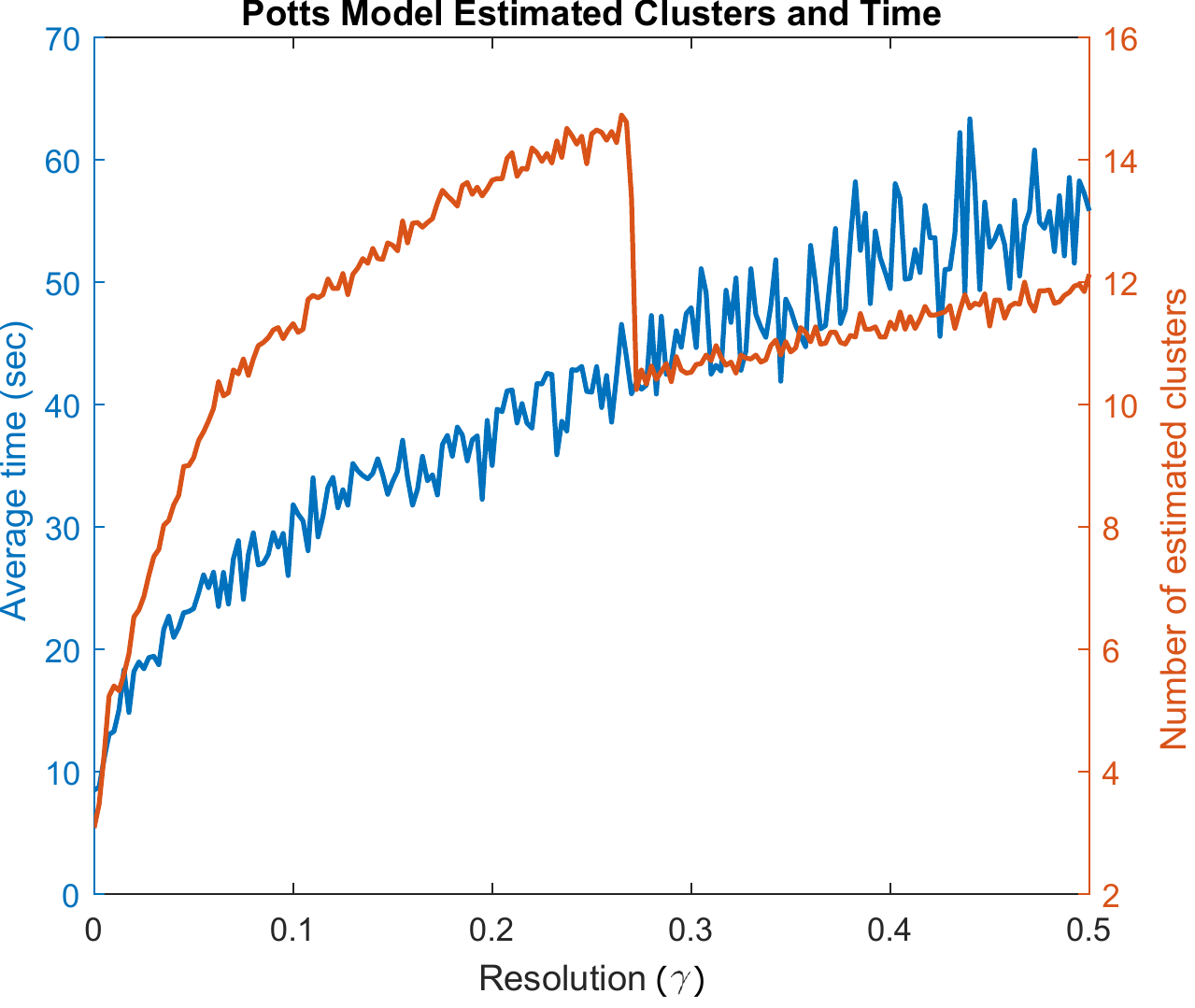}
  \caption{Average convergence time and number of estimated clusters as a function of Potts model resolution ($\gamma$), when clustering synthetic data (shown in \cref{fig:synthdata}). Results were averaged over all down-sampled lengths $100\leq M'' \leq600$. The algorithm determines the correct number of clusters ($=4$) at $\gamma\approx0.02$. The jump seen in the number of clusters at $\gamma \approx 0.275$ resembles those seen in previous work~\cite{JMI:JMI12097} and occurs at unstable resolutions.}
  \label{fig:timeandclusters}
\end{figure}

\begin{figure}
  \includegraphics[width=\linewidth]{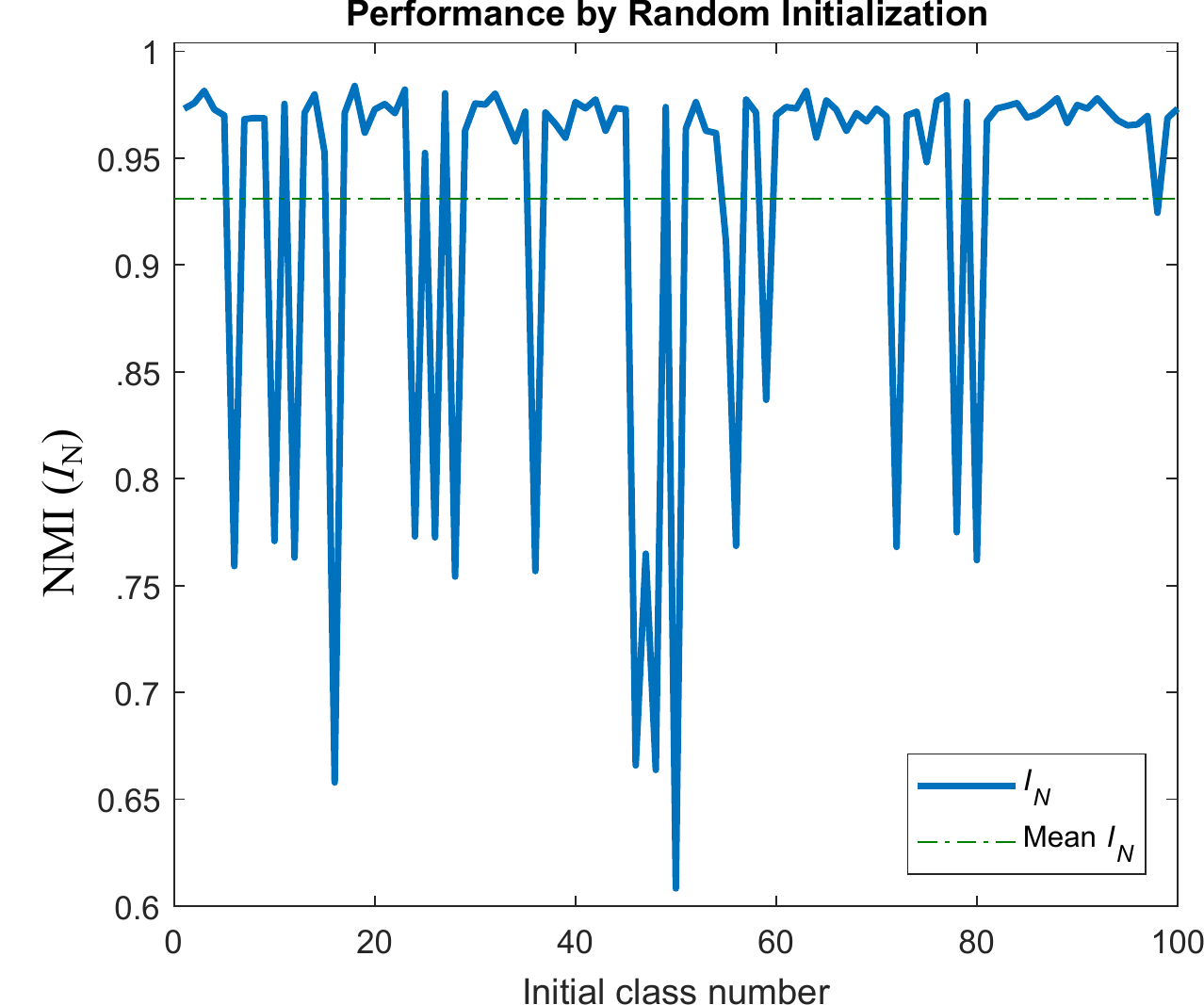}
  \caption{Potts model clustering performance as a function of the number of randomized initial classes, when clustering a synthetic dataset. Result was generated using $\gamma=0.02$ and $M''=250$. Overall we find no correlation between the number of random initialization classes and method performance, indicating that our algorithm is capable of converging to an optimal solution independent of initialization. We do observe random drops in performance, which we attribute to convergence to suboptimal local minima~\cite{JMI:JMI12097}.}
  \label{fig:initialclasses}
\end{figure}

For synthetic data clustering, the Potts model was allowed to discover the number of data clusters. The resolution, $\gamma$, was tuned to optimize clustering, and the maximum number of nodes, $M''$, was altered to study its effect on clustering performance, shown in \cref{fig:surfNMI}. We find that for this dataset, Potts model clustering performs best at $\gamma\approx 0.02$ (\cref{fig:NMIvsgamma}), and performance increased with increasing nodes $M''$. However, at $M'' \approx 300$, performance gains begin to have diminishing returns, as highlighted in \cref{fig:NMIvsphi}. To optimize clustering performance and time, $M''$ should be given as $\approx 350$. This is an acceptable compromise between method performance (\cref{fig:NMIvsphi}) and speed (\cref{fig:timevsnodes}). In practice, the average clustering times presented in \cref{fig:timevsnodes} would be significantly faster with proper resolution selection, as the clustering time increases with the number of classes determined by the algorithm (\cref{fig:timeandclusters}). Finally, to show the robustness of our method, we present our algorithm's performance as a function of the number of random initial classes. While the method occasionally suffers as a result of convergence to a sub-optimal local minima, \cref{fig:initialclasses} shows that the performance is consistent regardless of the initialization.

\subsubsection{Method Comparison}\label{method comparison}
\begin{figure}
  \includegraphics[width=\linewidth]{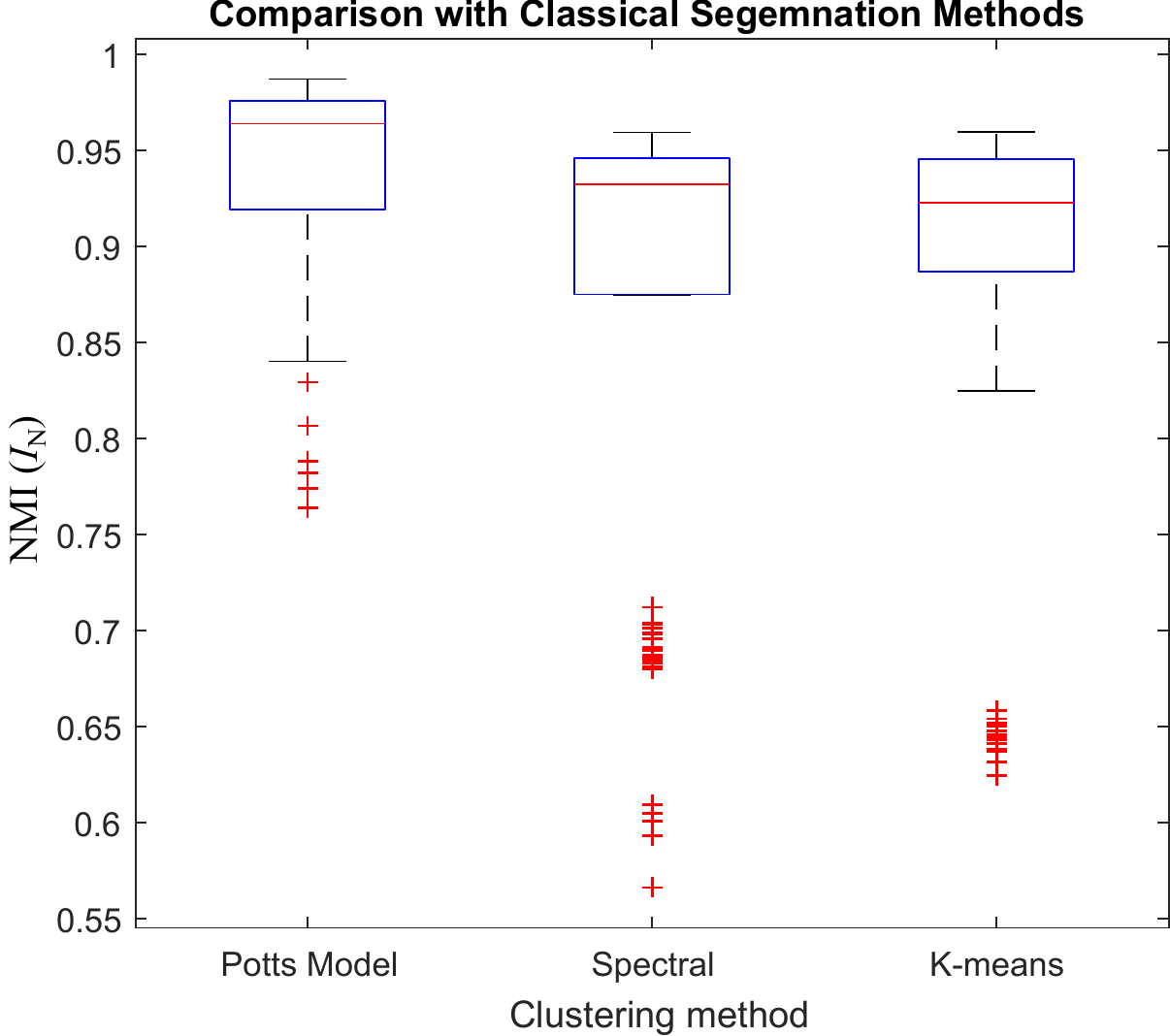}
  \caption{Potts model segmentation performance with respect to classical segmentation methods -- Spectral and K-means clustering. For spectral and K-means, the number of classes employed to investigate  was four. Potts model clustering was performed at $\gamma=0.02$ and $M''=350$. The Potts model outperforms the other two methods.}
  \label{fig:methodComp}
\end{figure}

\begin{figure}
  \includegraphics[width=\linewidth]{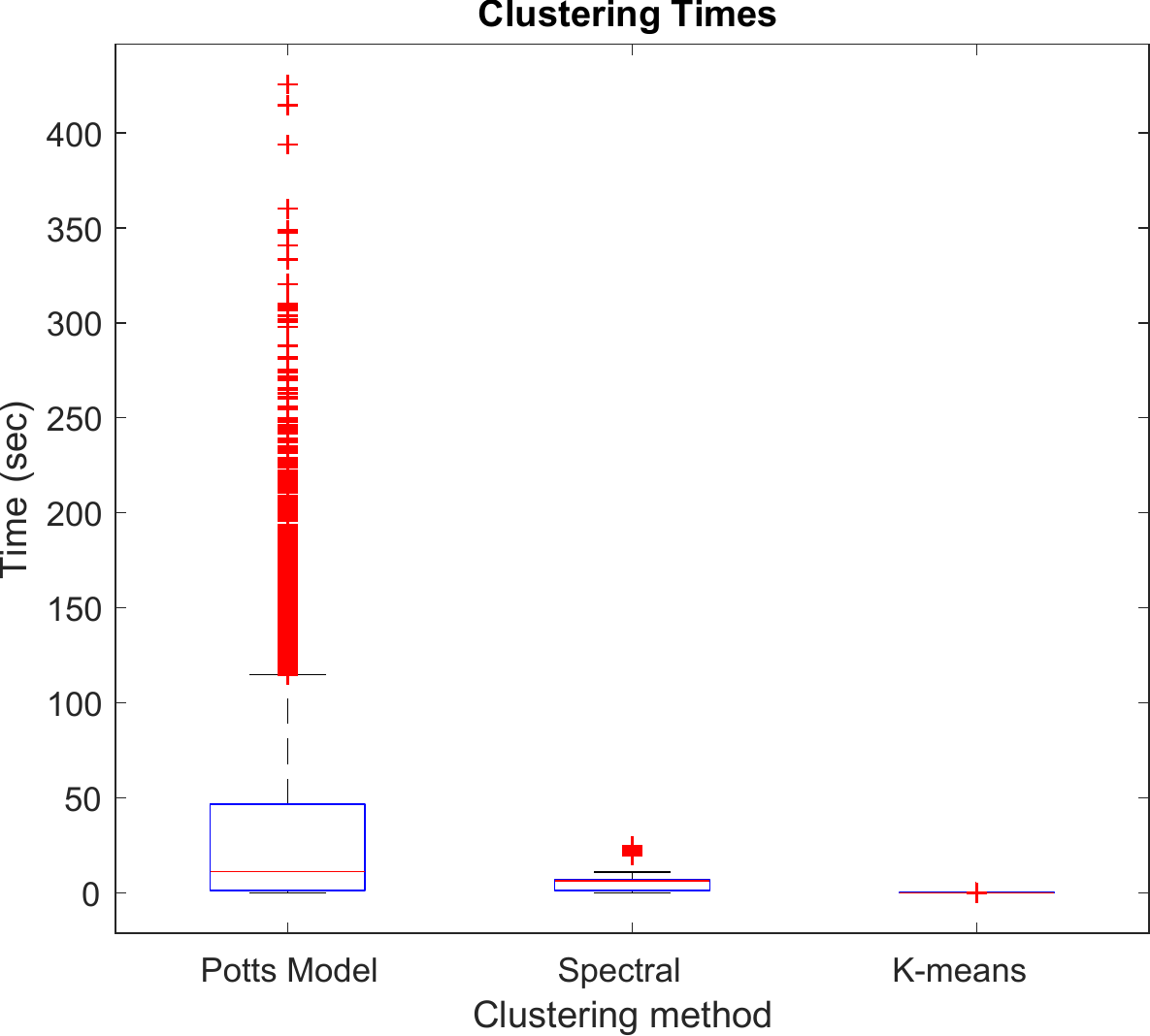}
  \caption{Clustering time by method. For Spectral and K-means, the number of classes employed to investigate was four. Potts model clustering was performed at $0\leq\gamma\leq0.5$ and $100\leq M'' \leq600$. This result is a statistical representation of 10000 realizations per method. Clustering using the Potts model is longer than the other methods, however, the Potts model result is likely to be skewed by the inclusion of clustering where $M'' > 350$.}
  \label{fig:methodTime}
\end{figure}

To compare the clustering performance, segmentation was performed on the generated synthetic data (\cref{fig:synthdata}) first using classical K-means~\cite{1017616} and spectral clustering~\cite{NIPS2001_2092, 6213172}. For both methods, the correct number of classes was specified. For our Potts model segmentation, the resolution parameter in Eq. (\ref{eq25}) was set to be the optimal value, $\gamma=0.02$. The results of 100 clustering realizations are presented in \cref{fig:methodComp}. The Potts model outperforms the classical K-means and spectral clustering, having the best mean and maximal NMI. Here the Potts mean NMI $\approx0.96$ matches the optimal one as shown in \cref{fig:NMIvsgamma} and \cref{fig:NMIvsphi}. Additionally, \cref{fig:methodTime} presents the computational time taken by each method when clustering synthetic data. For the Potts model, this figure depicts the computation times for all $0< \gamma \leq 0.5$ (using $\gamma$ intervals to be $0.0025$) and $100 \leq M'' \leq 600$ (using $M''$ intervals to be $20$). Each of these cases, as well as each of K-means and spectral clustering methods, was conducted for 10000 realizations. We observe that the Potts model has higher variation in convergence time than spectral and K-means clustering methods; the Potts model is also slower than these other two methods when convergence time distribution is studied. When we study the {\it average time} taken by the individual methods for the respective evaluations, we found that such average is comparable across all the three methods.

\begin{figure}
	\includegraphics[width=\linewidth]{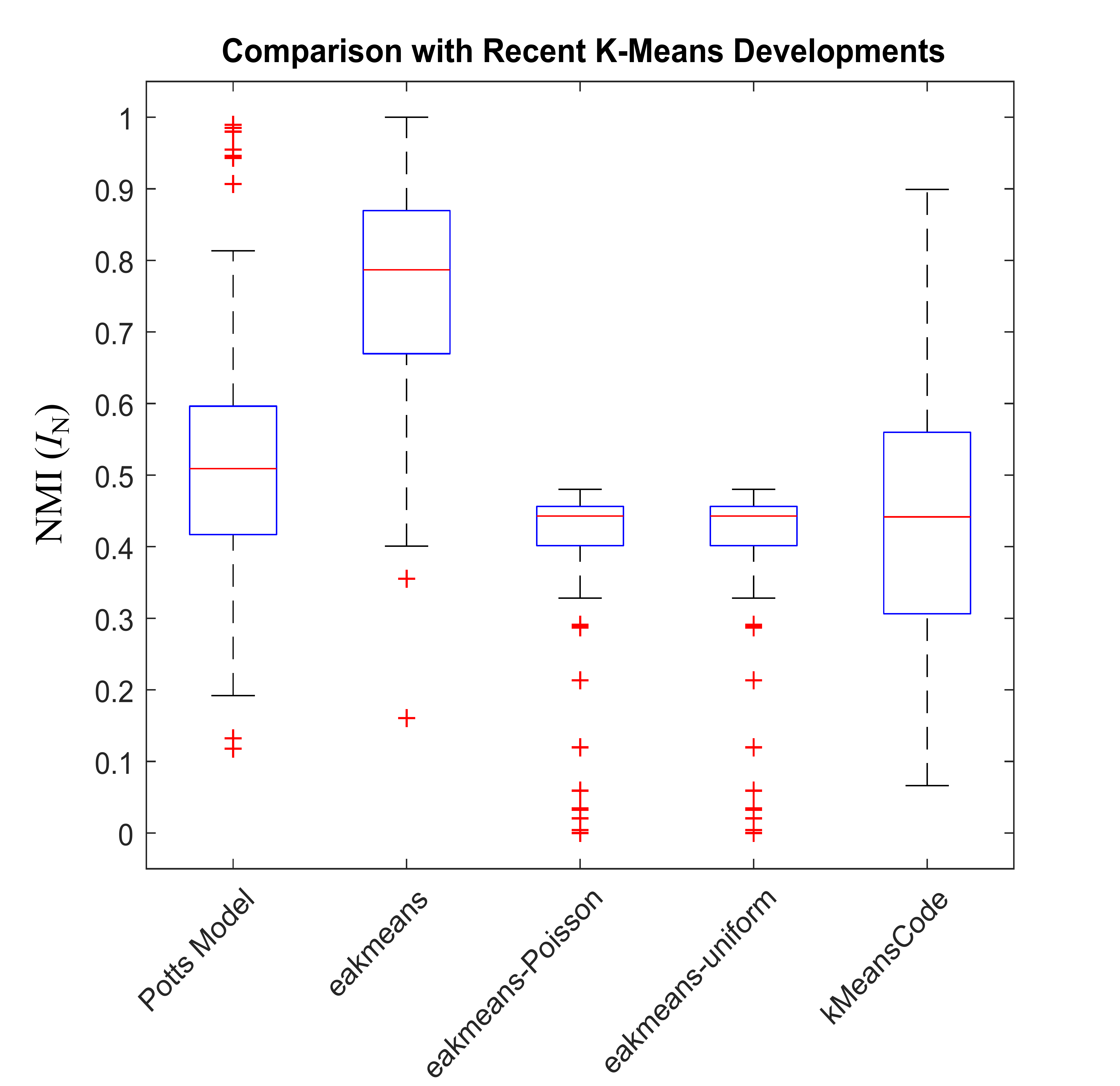}
  \caption{Method-wise clustering performance, evaluated on 100 realizations of 5000 synthetic data-points. Potts model segmentations were generated by selecting the lowest $\mathcal{H}$ for $0 < \gamma \leq 1$. Here eakmeans and kMeansCode represent methods described in Newling {\em et al.}~\cite{2016arXiv160202514N, 2016arXiv160202934N} and Shindler and Braverman {et al.}~\cite{Shindler:2011:FAK:2986459.2986724, Braverman:2011:SKM:2133036.2133039}, respectively. The segmentationâs for eakmeans and kMeansCode were generated using four classes. To fairly compare the Potts model's automatic model selection to K-means, performance of the eakmeans algorithm was evaluated for a non-specific number of classes; eakmeans-Poisson and eakmeans-uniform segmentations represent the average NMI for 100 realizations of eakmeans when the number of classes was randomly sampled from a Poisson and uniform distribution. These distributions were constructed of numbers ranging from 1 to 10, with the Poisson mean set to be the correct number of classes. The synthetic data-sets used in this comparison contain more intra-class variance to exemplify method performance.}
  \label{fig:clustercomp}
\end{figure}

Improvements to the K-means clustering algorithm have recently been proposed in the literature; see Newling {\em et al.}~\cite{2016arXiv160202514N, 2016arXiv160202934N}, Shindler {\em et al.}~\cite{Shindler:2011:FAK:2986459.2986724}, and Braverman {\em et al.}~\cite{Braverman:2011:SKM:2133036.2133039}. We evaluate our method against two such recent implementations of K-means; namely, against Newling {\em et al.}~\cite{2016arXiv160202514N} and Shindler {\em et al.}~\cite{Shindler:2011:FAK:2986459.2986724}. These methods address the primary issue of the classical K-means method which is computationally inefficient and does not scale well with increasing data size. These recent K-means algorithms, referred to by the authors in their original codes as ``eakmeans''~\cite{2016arXiv160202514N} and ``kMeansCode''~\cite{Shindler:2011:FAK:2986459.2986724} are compared to Potts model segmentation in \cref{fig:clustercomp}. We find that the eakmeans algorithm outperforms the Potts model segmentation when the correct number of data classes is specified. However, a fair comparison of the Potts model and eakmeans performance is challenging as the Potts model Hamiltonian cost performs automatic model selection~\cite{castle2010evaluating, PhysRevE.81.046114}. Namely, by tuning the resolution parameter $\gamma$ the Potts model method converges to final number of estimated clusters (see \cref{graph segmentation}). Lower values of $\gamma$ lead to a smaller number of segments while a higher value of $\gamma$ leads to higher number of segments. In this example, one such $\gamma$ value leads to the correct number of classes; however, what is also important is how much the estimated clusters for different $\gamma$ values overlap with the correct clusters. Therefore, in order to present a fair comparison, we computed the Potts model solution for several $\gamma$ values in a range $0 < \gamma \leq 1$ as well, and selected the solution corresponding to $\gamma$ with minimal $\mathcal{H}$. For an equal comparison, the number of clusters specified to the eakmeans algorithm is also randomly sampled from a Poisson (eakmeans-Poisson) and uniform distribution (eakmeans-uniform) respectively, drastically reducing the eakmeans method performance; see \cref{fig:clustercomp}. 

\subsection{Image Segmentation}\label{image segmentation}
We discuss below the performance of our proposed Potts model minimization for image segmentation task in segmenting Berkeley segmentation dataset~\cite{MartinFTM01}, as well as in segmenting renal glomerular compartments for quantitative assessment of renal biopsies.

\subsubsection{Benchmark Image Segmentation}\label{benchmark image segmentation}
\begin{figure}
  \includegraphics[width=\linewidth]{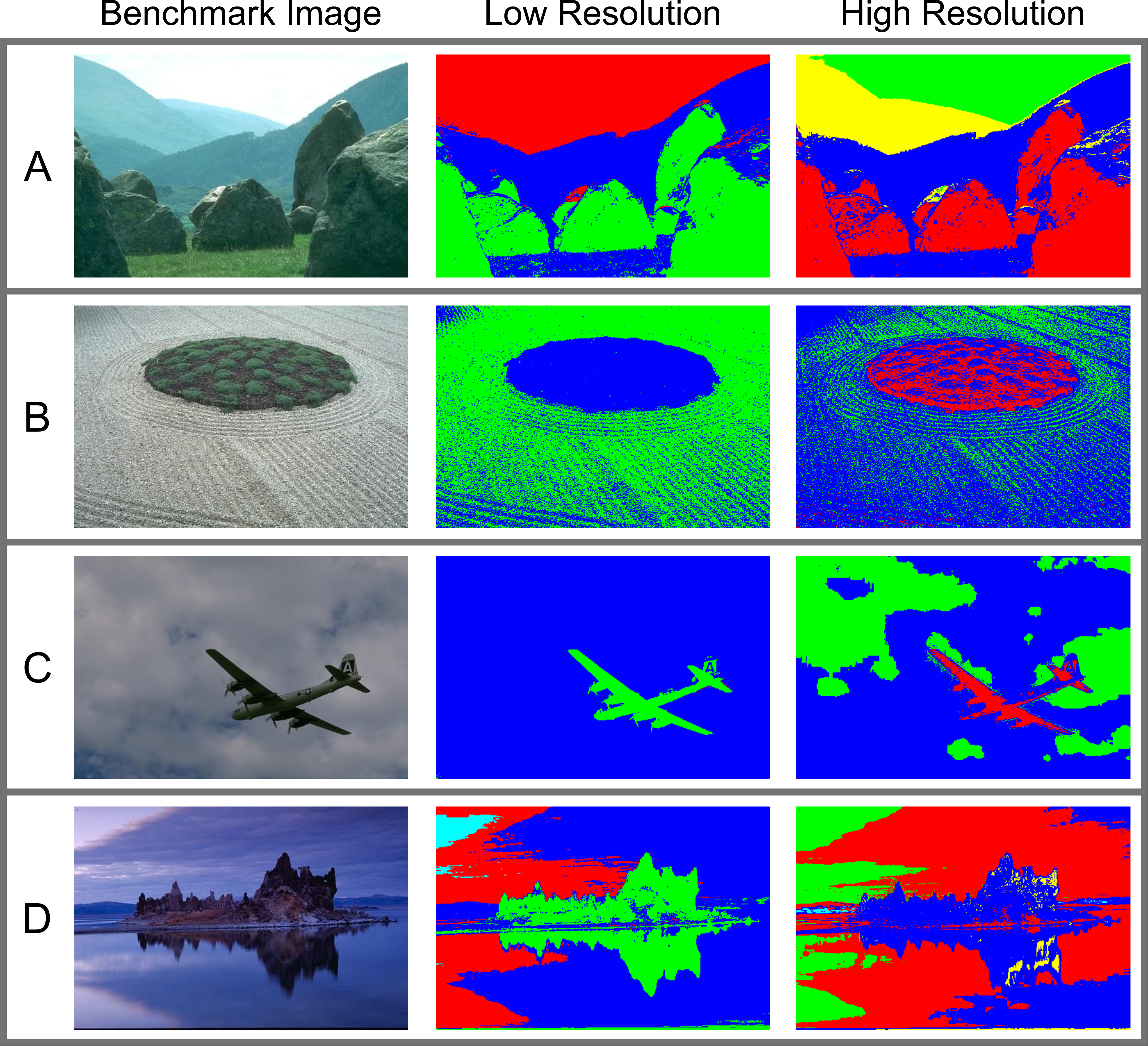}
  \caption{Benchmark image segmentation using the proposed algorithmic solution of the Potts model method. Segmentation at low resolution ($\gamma = 5$) and high resolution ($\gamma = 50$) were performed on four benchmark images. Raw pixel RGB values were used as image features, no pre-processing was done to enhance segmentation. Low resolution segmentations were completed in $\approx 4$ iterations, with high resolution longer to converge, $\approx 6$. Here color represents segments as determined by the algorithm. }
  \label{fig:benchmarkseg}
\end{figure}

To validate our method on an independent dataset, segmentation was performed on the Berkeley segmentation dataset~\cite{MartinFTM01}. The results of high/low resolution segmentation of four benchmark images are presented in \cref{fig:benchmarkseg}. We found that using $M'' = 300$ nodes gave good segmentation performance while optimizing algorithmic speed; taking an average of $3.87$ sec to segment each $481\times321$ pixel image. Quantitative evaluation using this dataset is limited, as our algorithm was given pixel RGB values as image features, however, it can be seen that higher resolution gives more specific segmentation. 

\subsubsection{Glomerular Segmentation}\label{glomerular segmentation}
To validate Potts model segmentation on an independent dataset, we used images of glomerular regions extracted from histologically stained whole slide murine renal tissue slices. All animal studies were performed in accordance with protocols approved by the University at Buffalo Animal Studies Committee. The glomerulus is the blood-filtering unit of the kidney; a normal healthy mouse kidney typically contains thousands of glomeruli~\cite{young_atlas, pollak_glom}. Basic glomerular compartments are Bowman's and luminal spaces, mesangial matrix, and resident cell nuclei~\cite{kriz1998progression}. In this paper, we demonstrate the feasibility of our proposed method in correctly segmenting these three biologically relevant glomerular compartments, see \cref{fig:pottsGlomSeg}. Image pixel resolution was $0.25$ $\upmu\mathrm{m}$ in this study. Once again we found that using $M''\approx 300$ nodes gave good segmentation performance, for segmenting $\approx499\times441$ pixel glomerular RGB image, while optimizing algorithmic speed. 

\begin{figure}[H]
	\includegraphics[width = \linewidth]{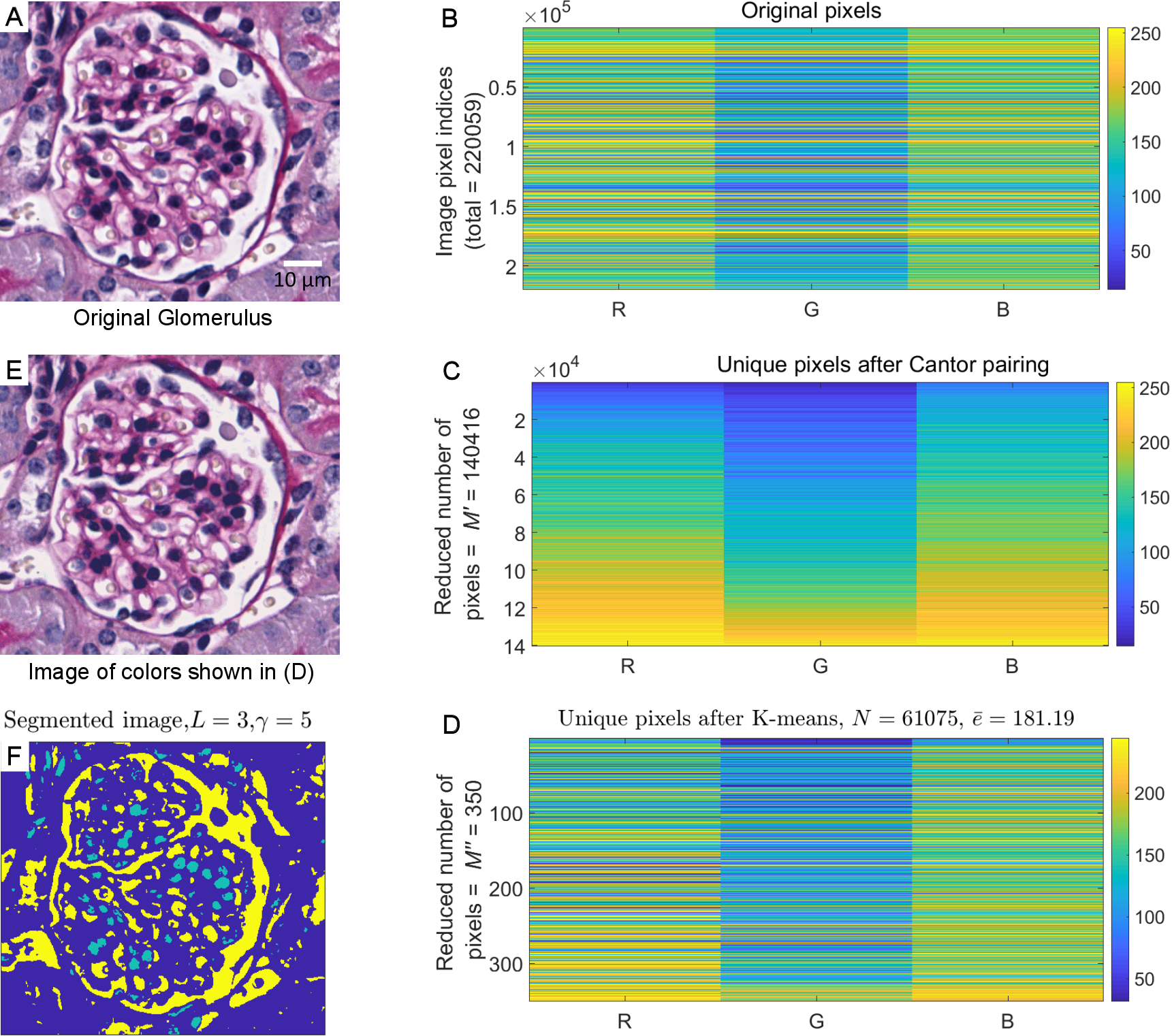}
	\caption{ Steps of the proposed segmentation. (A) A histologically stained glomerulus image, containing $499 \times 441$ RGB pixels. (B) Vectored form of original 8-bit image pixels. (C) Vectored form of image pixels after Cantor pairing with $M' = 140416$ distinct colors. (D) Vectored form of the same set of image pixels after K-means based down-sampling with $M'' = 350$ colors. Number of edges for the full graph using the rows shown here as nodes is $N = 61075$ and mean edge strength $\bar{e} = 181.19$. (E) Image formed using the colors shown in D at the original image size. The structural similarity index~\cite{li2010content, 1284395} between the image in A and E is $0.97$, despite extensive reduction in color information. (F) Segmented image using the full graph formed using the rows shown in D as nodes. Segmentation was conducted with $\gamma = 5$, and the total numbers of segments obtained is $L=3$. }
	\label{fig:gloms}
\end{figure}

\cref{fig:gloms} shows the steps of the proposed segmentation using actual numbers used. Here we show a histologically stained glomerulus image, containing $499 \times 441$ RGB pixels, and the vectored form of this original 8-bit image pixels corresponding to \cref{eq1}. Vectored form of image pixels after Cantor pairing has rows $M' = 140416$ distinct colors corresponding to \cref{eq8}. Vectored form of the same set of image pixels after K-means based down-sampling  has rows $M'' = 350$ colors; see discussion on this method at the second half of \cref{dataset definition}. Number of edges for the full graph using the down-sampled data is $N = 61075$ and mean edge strength $\bar{e} = 181.19$. Structural similarity index~\cite{li2010content, 1284395} between the image formed using the reduced colors at the original image dimension and the original glomerulus image is $0.97$, despite extensive reduction in color information. Segmentation was conducted using the full graph formed using the down-sampled data with $\gamma = 5$; see \cref{graph segmentation} for a discussion on the method. The total numbers of segments was obtained to be $L=3$.

\begin{figure}
  \includegraphics[width=\linewidth]{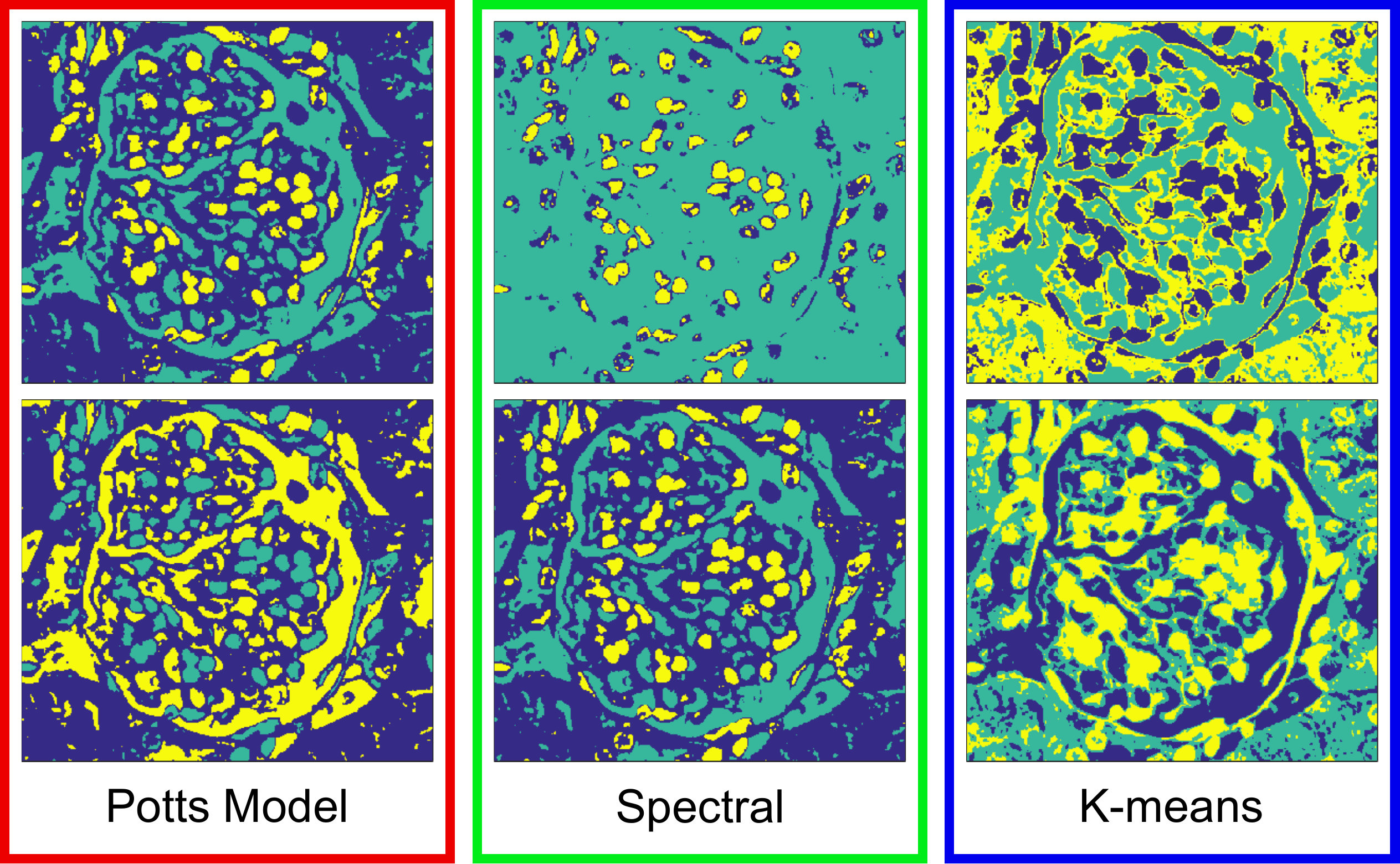}
  \caption{Glomerular compartment segmentation using different segmentation methods. The original image is depicted in \cref{fig:gloms}-A. Identical parameters were used to generate both segmentations for each method: Potts model resolution was $\gamma= 1$, spectral and K-means employed three classes for the investigation. For all segmentation, pixel RGB values were used as the three image features. For potts model segmentation, $M''=350$ was used, as depicted in \cref{fig:gloms}-E.}
  \label{fig:allGlomSeg}
\end{figure}

Potts model segmentation was also compared against spectral~\cite{NIPS2001_2092, 6213172} and classical K-means clustering~\cite{1017616} based segmentation; see \cref{fig:allGlomSeg}. Two different realizations of the segmentation using identical parameters are represented for each method. Qualitatively, we found the Potts model to give the best, and most reproducible segmentation. Spectral clustering also performs well, but gives less reproducible segmentation, while K-means does a poor job distinguishing glomerular compartments. Potts model determines the three image classes using the baseline resolution ($\gamma=1$), resembling the three biological compartments depicted in \cref{fig:pottsGlomSeg}.

We analyzed the performance of our three basic glomerular compartment (Bowman's and luminal space, mesangial space, and resident cell nuclei) segmentation as stated above using renal tissue histology images from three wild-type mice~\cite{NEP:NEP796}. Testing was done on five glomerular images per mouse, and evaluated against ground-truth segments generated by renal pathologist Dr.\ John E.\ Tomaszewski (University at Buffalo). The performance of our method was compared against compartmental segments jointly generated by two junior pathologists' (Dr.\ Buer Song and Dr.\ Rabi Yacoub) manual segmentations. For each compartment we computed precision and accuracy of automated and manual segmentations per glomerulus. The respective metrics were averaged for all the compartments per glomerulus for each of the automated and manual segmentations. The respective averages were divided by the respective times taken by manual and automated segmentations. \cref{fig:pathcomp} depicts the resulting precision and accuracy~\cite{fletcher_pa} per time. Average precision and accuracy per unit segmentaion time (automatic or manual) were computed across mice, and standard deviations of these metrics over mice were computed. Comparison indicates Potts model segmentation significantly outperforms manual annotation with high efficiency.

\subsubsection{Comparison with Storath {\em et al.}~\cite{storath2014fast} and Stutz {\em et al.}~\cite{stutz2018superpixels}}\label{storath}
\begin{figure}
	\includegraphics[width=\linewidth]{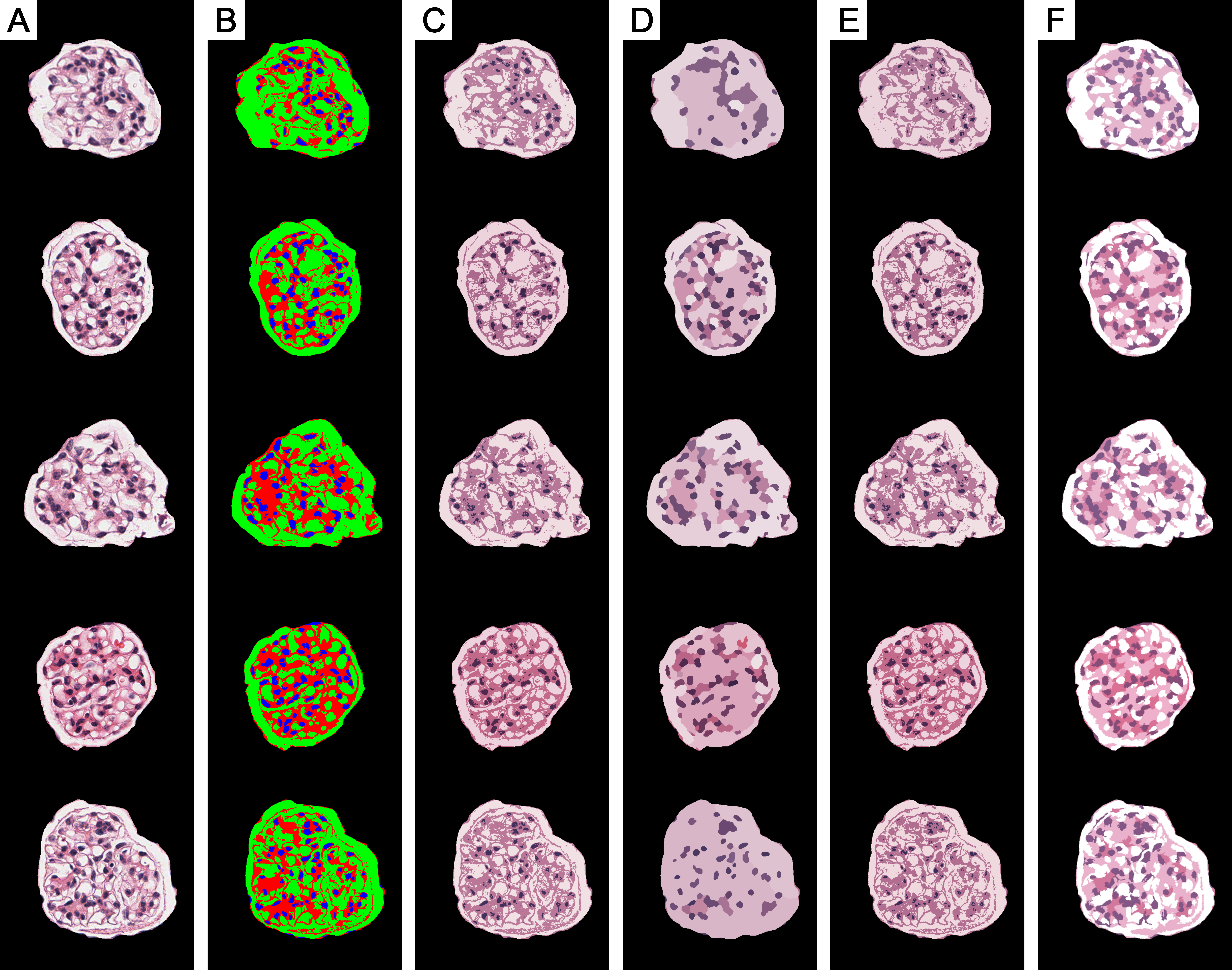}
	\caption{Proposed Potts model segmentation performance comparison with dynamic programming mediated optimization based Potts model segmentation method proposed by Storath {\em et al.}~\cite{storath2014fast} and a Markov random field method which uses superpixels for speed gains by Stutz {\em et al.}~\cite{stutz2018superpixels}. (A) Histologically stained five glomerular images of normal control mice kidney tissue sections. (B) Ground-truth segmentation of nuclei, Bowman's and luminal space, and mesangial matrix regions. (C) Our proposed Potts model segmentation performance. (D) Segmentations by the method proposed by Storath {\em et al.}~\cite{storath2014fast}. (E) Segmentation by our proposed method using initial segments obtained via segmentations obtained by the method of Storath {\em et al.}~\cite{storath2014fast}. (F) Segmentation by the Markov random field and superpixel method of Stutz {\em et al.}~\cite{stutz2018superpixels}.}
	\label{fig:lutni17}
\end{figure}

\cref{fig:lutni17} depicts the performance of  our proposed Potts model segmentation with dynamic programming mediated optimization based Potts model segmentation method proposed by Storath {\em et al.}~\cite{storath2014fast}, and the Markov random field with superpixels method of Stutz {\em et al.}~\cite{stutz2018superpixels}. We used histologically stained five glomerulai images of normal control mice kidney tissue sections, and attempted to segment nuclei (dark region), Bowman's and luminal spaces (gray region), and mesangial matrix (pink region). Ground-truth segmentation of these compartments was done by the first author Mr.\ Brendon Lutnick under the supervision of renal pathologist Dr.\ John E.\ Tomaszewski. We considered three cases for the segmentation. Namely, segmentations were performed using our method, methods by Storath {\em et al.} and Stutz {\em et al.}~\cite{stutz2018superpixels}, and a combined method initialized by the output of the method by Storath {\em et al.} with final segmentation conducted by our method. We compared the computationally multi-class segmented image with the ground-truth using normalized mutual information (NMI) defined in \cref{eq33}. Our method outperformed the method by Storath {\em et al.} and Stutz {\em et al.}~\cite{stutz2018superpixels} with 2.6X and 1.7X better performance based on NMI performance metric respectivly, while showing 2.33X slower speed in convergence. The combined method requires significantly higher time to converge. This is because the Storath {\em et al.} method based initialization defines poor initialization of our method requiring more time to converge.

\subsection{Data Sharing for Reproducibility}\label{datshreprod}
All of the source code and images used to derive the results presented within this manuscript are made freely available from \url{https://goo.gl/V3NatP}. 

\section{Discussion}\label{discussion}
The primary intention of this paper is to provide an overview of our proposed method for Potts model based segmentation, where the results above are merely applications to validate our method when applied to specific segmentation tasks. These analyses were performed on the raw data, with no pre-processing enhancements or feature selection; as a result the image segmentation as presented in \cref{glomerular segmentation,synthetic data segmentation} may be sub-optimal. We expect image segmentation to be limited without the use of high level contextual features, or image pre-processing. However these examples help evaluate the computational performance and scalability of our method. For future applications in image segmentation, we will apply automated methods for feature selection such as sparse auto-encoders to represent image data in more meaningful dimensions \cite{ng2011sparse, 7163353}. The synthetic data presented in \cref{fig:synthdata} is more representative of actual separable data containing abstract features, and while our method provides an approximation to the optimal segmentation, it outperforms both standard K-means and spectral clustering, as well as a modern implementation of K-means~\cite{2016arXiv160202514N, 2016arXiv160202934N}; see \cref{fig:methodComp,fig:methodTime,fig:clustercomp,fig:allGlomSeg}.

To boost the algorithmic performance, we utilize several statistical assumptions, which reduce the complexity of computationally intensive problems. Namely, the inclusion of the Cantor pairing based and modified K-means down-sampling in \cref{dataset definition}. Here we propose $\alpha$, a parameter which broadens the criteria for convergence of the modified K-means iteration. Practically we have set $\alpha = 0.95$ to ensure that the number of nodes selected, $M''$, is within $95 \%$ of the user specified value $K$. We have found that this encourages fast convergence of the K-means while maintaining an acceptable level of accuracy. Additionally in \cref{eq18} we define the PID tuning parameters $K_\mathrm{p}$, $K_\mathrm{i}$, and $K_\mathrm{d}$ which have been assigned $0.5$, $0.05$, and $0.15$, respectively. We find that these values provide fast settling times, while minimizing overshoot, satisfying the length of the reduced dataset $\tilde{\mathbf{D}}'$ to be at least $\alpha K.$ Likewise, to calculate the average edge $\xbar{e}$ of the full graph model of the image using the approximated graph formed using $\tilde{\mathbf{D}}'$, we propose $M'''$, a parameter which determines the maximum data-points used in the calculation of $\xbar{e}$. Practically we define $M''' = 5000$, ensuring the calculation of $\xbar{e}$ is fast. We have found that using $M''' = 5000$ gives accurate and reliable estimation of $\xbar{e}$. Computing $\xbar{e}$ with the down-sampling resulted in $\approx 1 \%$ error, while exponentially increasing algorithmic speed.

\begin{figure}
	\includegraphics[width=\linewidth]{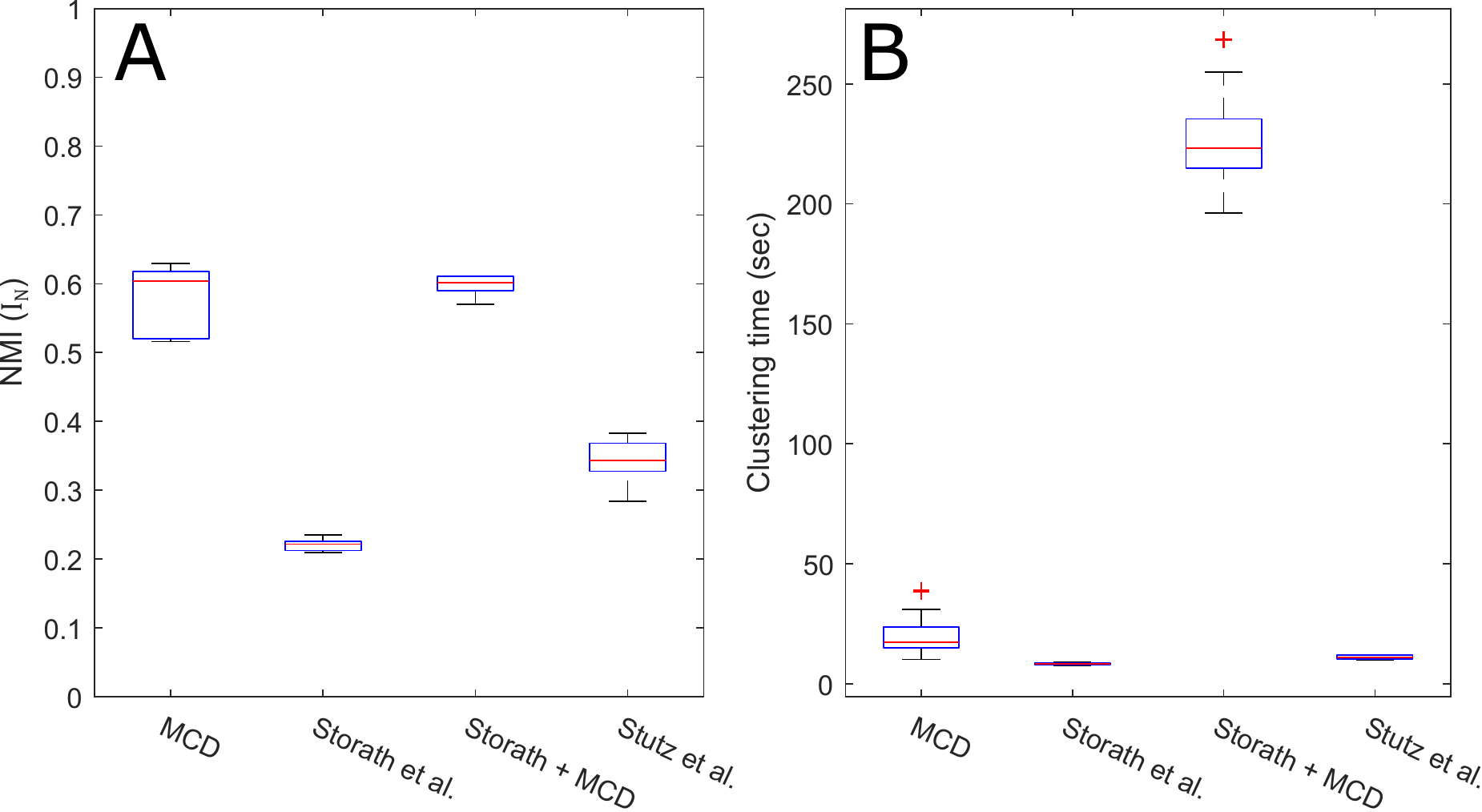}
	\caption{Quantitative performance comparison between our proposed Potts model segmentation, the method proposed by Storath {\em et al.}~\cite{storath2014fast} and the Markov random field with superpixel approach by Stutz {\em et al.}~\cite{stutz2018superpixels} in segmenting glomerular micro-compartments as shown in \cref{fig:lutni17}. Statistics was obtained using the five glomeruli images. Segmentations were performed using our method, method by Storath {\em et al.}, and a combined method initialized by the output of the method by Storath {\em et al.} with final segmentation conducted by our method. (A) Normalized mutual information (NMI) defined in \cref{eq33} is compared for all the cases, and (B) the respective segmentation times are compared as well. Our method offers 2.6X and 1.7X better performance based on NMI performance metric than Storath {\em et al.} and Stutz {\em et al.} respectively, while showing 2.33X slower speed in convergence. The combined method requires significantly higher time to converge.}
	\label{fig:lutni18}
\end{figure}

The algorithmic solution for Potts model minimization we present quickly converges to stable solutions, but is not immune to poor initialization. While the algorithm automatically determines the correct number of segments, poor initializations often converge to sub-optimal local minima. Practically this occurs when no energetically favorable move exists for any node in the system while optimizing the Potts model energy, there may be a better solution, but to find it would require moves that increase the Hamiltonian cost. The effects of poor initialization are presented in \cref{fig:initialclasses}, where the number of initial classes has no discernible trend on segmentation performance, but poor initialization likely leads to occasional performance loss. We found similar trend when we initialized our proposed method using the output of the method proposed by Storath {\em et al.}; see a discussion in \cref{storath,fig:lutni17,fig:lutni18}. The simplest solution is to repeat the Potts model minimization based segmentation several times, selecting the one with the lowest cost, $\mathcal{H}$. Similar strategy we adopted in our previous works~\cite{JMI:JMI12097, doi:10.1117/12.2036875, PhysRevE.81.046114, PhysRevE.85.016101}. Alternatively, future study of optimal initialization techniques could help discover a computationally easier work around. The effects of the resolution parameter $\gamma$ in Potts model minimization are not yet fully understood and in future work we plan to develop a theoretical framework for these effect through empirically study. Additionally we plan to study the effects of system perturbations on Hamiltonian minimization. Addition of robust perturbation functions to disturb system equilibrium, will likely benefit image segmentation and data clustering performance.

\section{Conclusion}\label{conclusion}
The Potts model provides a unique approach to large scale data mining, its tunable resolution parameter indirectly conducts automatic model selection and thus provides useful tools for segment discovery. Unlike other unsupervised approaches, the Potts model minimization allows us to determine the number of segments by leveraging the data structure and features. Previous work on the Potts model was limited due to inefficient algorithmic optimization of the Hamiltonian cost function~\cite{JMI:JMI12097, doi:10.1117/12.2036875, PhysRevE.81.046114, PhysRevE.85.016101}. Our approach circumvents this problem by utilizing statistical simplifications of input data, and offering an innovative iterative solution considering pixel relationship to segments during the iteration. The proposed data down-sampling serves to approximate the optimal solution as segmentation of large image dataset would be unfeasible without such assumptions. Our method inherently scales with the feature space of the data, specifically with the number of distinct pixel data-points. In practice, the resolution ($\gamma$) and down-sampling ($M''$) can be optimized via alternative projection to optimize segmentation and speed, respectively, allowing the use of our method for data mining and discovery on any dataset with a discrete feature set.


%

\section*{Disclosures}
The authors have no financial interests or potential conflicts of interest to disclose. 

\section*{Acknowledgment}
\begin{figure}
  \includegraphics[width=\linewidth]{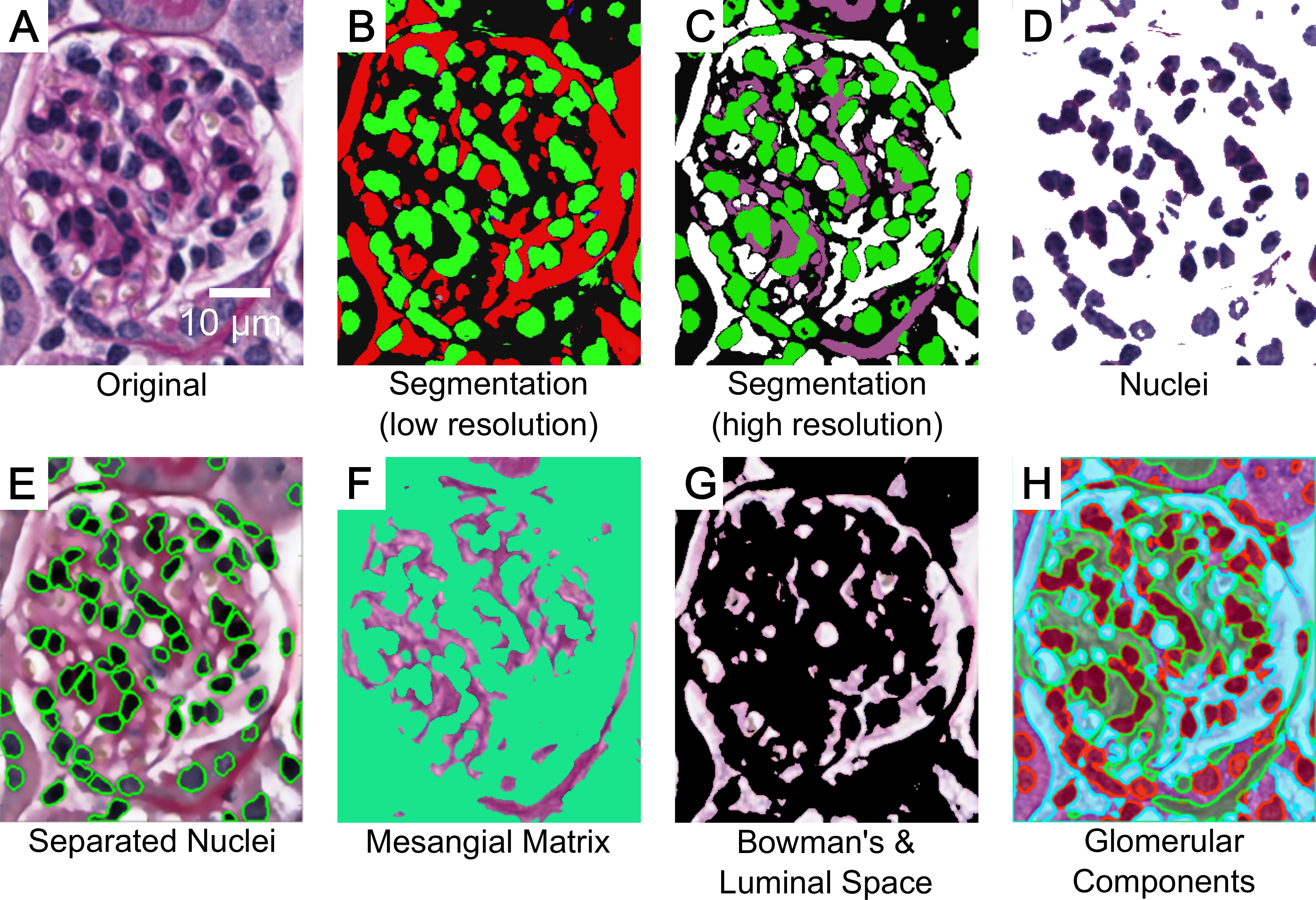}
  \caption{Murine renal glomerular compartment segmentation using a Potts model Hamiltonian. (A) The original glomerulus image, (B) low resolution ($\gamma\approx 0.5$) segmentation, and (C) high resolution ($\gamma\approx 5$) segmentation, (D) segmented nuclei, (E) separated nuclei superimposed on (A) using morphological processing, (F) segmented mesangial matrix, and (G) segmented Bowman's/luminal space. Compartment segments depicted in (D, F, and G) were obtained at optimally chosen $\gamma$ values where the respective compartment segmentations were verified by a renal pathologist (Dr.\ John E. Tomaszewski, University at Buffalo). (H) All three segmented components (D, F, and G) overlaying on the original image. All segmentations use $M'' = 350$ nodes. Here, color is used to signify segments, but is not conserved between panels, and the background colors in (D, F, and G) were chosen to enhance contrast in the image for visualization.}	
  \label{fig:pottsGlomSeg}
\end{figure}

\begin{figure}
	\includegraphics[width=\linewidth]{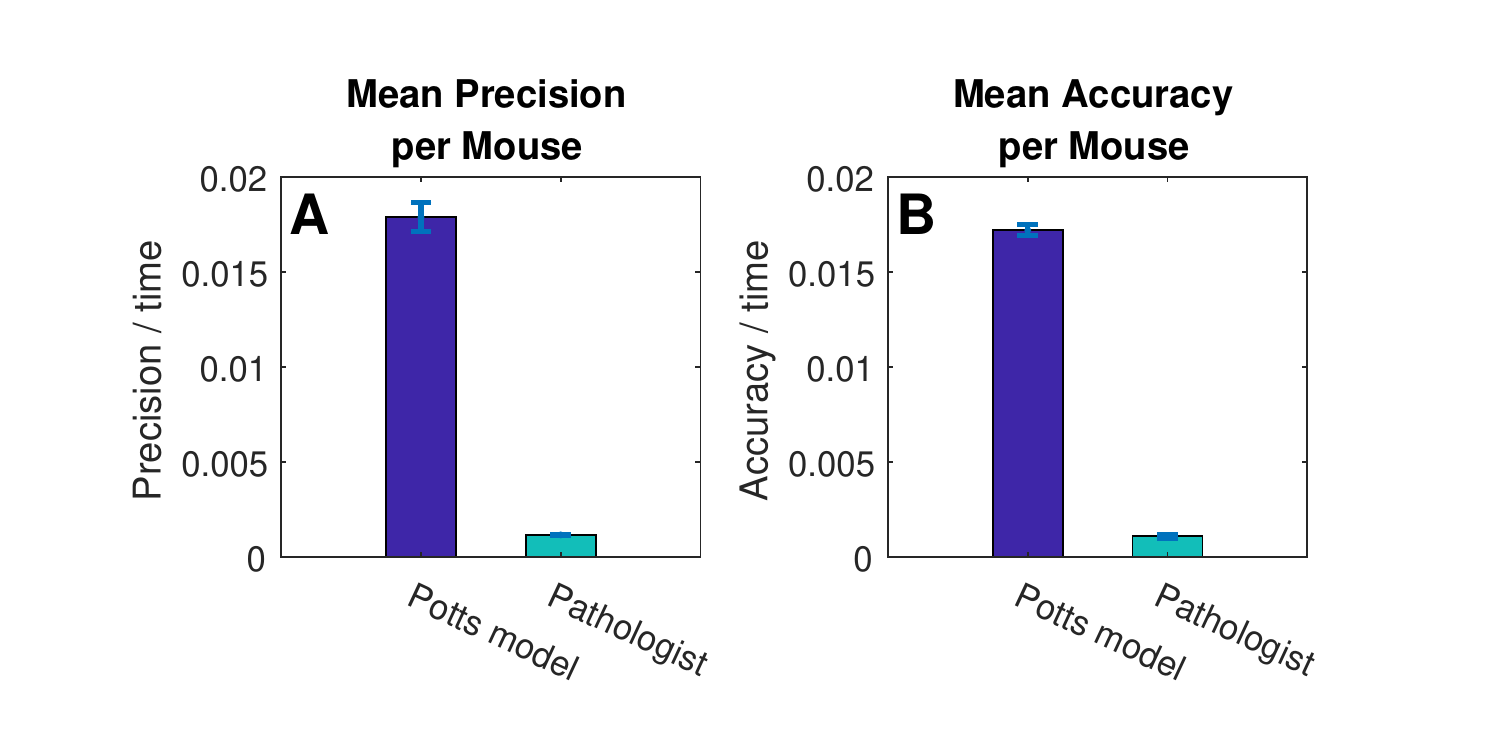}
	\caption{Comparison of performance between Potts model and manual methods in segmenting murine intra-glomerular compartments. (A) Precision per time. (B) Accuracy per time. Five glomeruli images per mouse from three normal healthy mice were used. Error-bars for the precision and accuracy metrics indicate standard deviation. Potts model based segmentation significantly outperforms manual method.}
	\label{fig:pathcomp}
\end{figure}

The project was supported by the faculty startup funds from the Jacobs School of Medicine and Biomedical Sciences, University at Buffalo, the University at Buffalo IMPACT award, NIDDK Diabetic Complications Consortium grant DK076169, NIDDK grant R01 DK114485, and NSF grant 1411229. We thank Dr.\ John E.\ Tomaszewski, Dr.\ Rabi Yacoub, and Dr. Buer Song for the pathological annotations (see \cref{fig:pottsGlomSeg} and \cref{fig:pathcomp}) and biological guidance they provided which was essential to the glomerular segmentation results presented in \cref{glomerular segmentation,storath}.




\newpage
\bibliographystyle{IEEEtran}
\bibliography{MCD_IEEE}

\newpage

\newpage

\newpage

\newpage

\newpage

\newpage

\newpage

\newpage

\newpage

\newpage

\newpage

\newpage

\newpage

\newpage

\clearpage

\newpage

\newpage

\newpage

\newpage

\end{document}